\documentclass{article}

    \PassOptionsToPackage{numbers, compress}{natbib}

\usepackage{microtype}
\usepackage{graphicx}
\usepackage{subcaption}
\usepackage{booktabs} 
\usepackage{tabularx, makecell}
\usepackage{multirow}
\usepackage{float}
\usepackage{amsmath}
\usepackage{wrapfig}
\usepackage[preprint]{neurips_2026}
\usepackage[table,xcdraw]{xcolor}
\usepackage[utf8]{inputenc} 
\usepackage[T1]{fontenc}    
\usepackage{hyperref}       
\usepackage{url}            
\usepackage{booktabs}       
\usepackage{amsfonts}       
\usepackage{nicefrac}       
\usepackage{microtype}      
\usepackage{xcolor}         
\definecolor{rowlightblue}{RGB}{223,243,250}

\title{GROW: Aligning GRPO with State-Action Modeling for Open-World VLM Agents}

\author{
    Xiongbin Wu$^{1,2}$ \quad Zhihao Luo$^{2,3}$ \quad Shanzhe Lei$^{2}$ \quad Lechao Zhang$^{2,3}$ \quad Xuhong Wang$^{2}$ \AND 
    JIE YANG$^{1}$ \quad Zhonglong Zheng$^{4}$ \quad Yuanjie Zheng$^{5}$ \quad Xin Tan$^{2,3}$\thanks{Corresponding author} \quad Wei Liu$^{1}$\thanks{Corresponding author} \\
    \\
    $^1$Shanghai Jiao Tong University \quad $^2$Shanghai Artificial Intelligence Laboratory \\
    $^3$East China Normal University \quad $^4$Zhejiang Normal University \quad $^5$Shandong Normal University
}

\begin{document}
\maketitle
\begin{abstract}
Recently, vision-language model (VLM) agents have shown promising progress in open-world tasks, where successful task completion often requires multiple turns of visual perception and action execution. 
However, existing methods still rely primarily on Supervised Fine-Tuning (SFT) with expert demonstrations, while the advanced reinforcement learning (RL) algorithm, specifically Group Relative Policy Optimization (GRPO), has not been effectively employed for multi-turn RL in these tasks because standard GRPO requires full trajectories as training samples which leads to excessively long context and noise.
To address this issue, we propose GROW, a RL framework for open-world VLM agents that decomposes collected trajectories into state-action samples, and computes advantages between these samples rather than treating a full trajectory as a single entity.
We further provide a surrogate analysis indicating that, even though the grouped samples are conditioned on different local states rather than an identical prompt context, the objective can preserve the core relative policy optimization signal of GRPO under simplifying assumptions. 
Experiments on more than 800 Minecraft tasks show that our method achieves state-of-the-art (SOTA) performance, demonstrating the effectiveness of our proposed RL framework for open-world VLM agents.
\end{abstract}

\section{Introduction}

VLM agents have become increasingly capable in open-world environments~\citep{tan2025lumineopenrecipebuilding, magne2026nitrogenopenfoundationmodel, li-etal-2025-jarvis, zhou2026main}. 
In these domains, an agent must repeatedly interpret visual states, choose actions, and adapt its behavior through multi-turn interaction with the environment. 
Most recent efforts improve such agents through SFT on expert demonstrations~\citep{Ouyang2026GameWorldTS, wang2025game}, enabling them to imitate task-relevant perception-action behaviors from curated demonstrations.
However, SFT relies on large amounts of high-quality expert data, whose collection is often expensive and difficult to scale. 
Moreover, prior studies~\citep{shi2025mobileguirladvancingmobilegui,ye2025etpr1evolvingtopologicalplanning,zhang2025activevlnactiveexplorationmultiturn} have shown that SFT alone can lag behind RL-trained VLM agents in performance.
These limitations motivate the need for advanced RL methods that can effectively train open-world VLM agents through interaction.

\begin{wrapfigure}{r}{0.48\textwidth} 
    \centering
    \vspace{-2mm} 
    \includegraphics[width=0.48\textwidth]{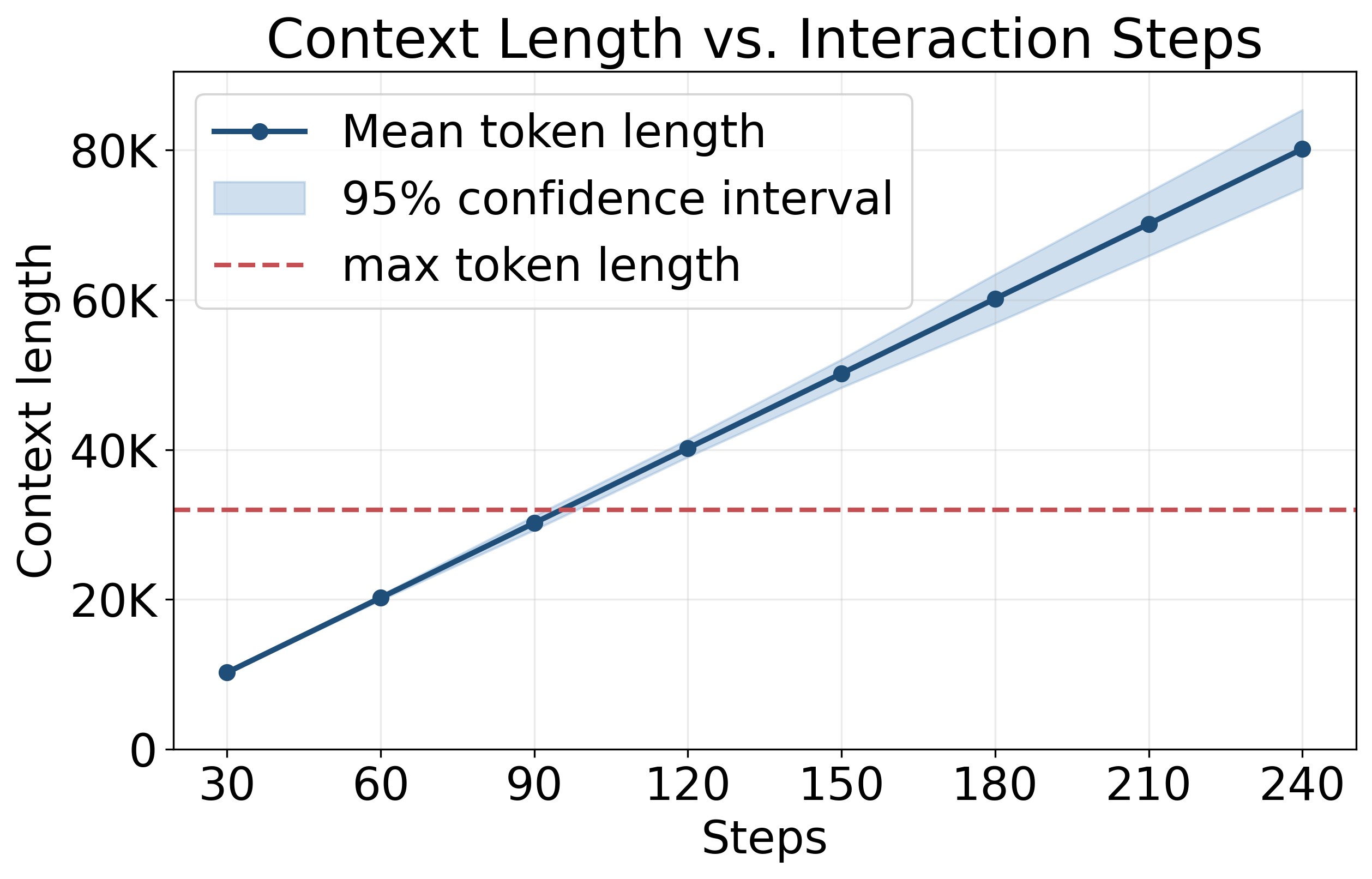} 
    \caption{Context length in the trajectory increases with the number of interaction steps between the VLM agent and the environment, often exceeding the maximum token length as interactions accumulate.}
    \label{fig:steps}
\end{wrapfigure}

Recent researchs on VLM~\citep{luo2025navimaster, li2026compassnav, hu2026longnavr1horizonadaptivemultiturnrl, zhang2025activevlnactiveexplorationmultiturn, ye2025etpr1evolvingtopologicalplanning, shi2025mobileguirladvancingmobilegui} which use GRPO~\citep{shao2024deepseekmath} as their RL algorithms have shown the effectiveness of GRPO in improving foundation-model policies through group-wise relative optimization. 
By comparing multiple sampled outputs within the same group, GRPO constructs relative advantages without training an additional value model, making it especially suitable for large-scale VLM optimization where value estimation can be costly and unstable. 
These strengths make GRPO a natural algorithmic basis for refining VLM agents through environment interaction. 
However, directly transferring GRPO to open-world tasks is nontrivial.
Standard GRPO compute advantages across trajectories conditioned on the same prompts and optimize these full trajectories. As shown in Figure~\ref{fig:steps}, it can introduce excessively long context and also include too much noise in the context when full trajectories are used to predict actions in open-world tasks. 

To address this issue, we propose \underline{G}\underline{R}PO for \underline{o}pen-\underline{w}orld VLM agents (GROW), a RL framework that adapting GRPO to open-world tasks where trajectories are often too long and agents make decisions often based on short-horizon states. 
GROW first applies trajectory decomposition to collected rollouts trajectories and then compute relative advantages among the state-action samples within the same rollout groups. 
However, this reformulation also introduces a theoretical issue that does not arise in standard GRPO. 
After trajectory decomposition, grouped samples are no longer different responses to the same prompt. 
At this, we provide a surrogate analysis under simplifying assumptions shows that the proposed objective can still provide an effective relative policy optimization signal.

Our main contributions are summarized as follows:
\begin{itemize}
    \setlength{\parskip}{0pt}

    \item We propose GROW, a RL framework for open-world VLM agents. 
    The framework combines cold-start training on state-action samples with GRPO-based policy refinement through trajectory decomposition.

    \item We provide a surrogate analysis suggesting that, under reasonable tractability approximations, relative policy optimization remains effective in GROW even when grouped samples are conditioned on different local states rather than an identical prompt context.

    \item We instantiate and test the GROW mainly in Minecraft~\citep{lin2023mcu, li-etal-2025-jarvis, bai2025qwen25vltechnicalreport}.
Across more than 800 Minecraft tasks,
ranging from embodied spatial navigation and precise GUI manipulation to highly dynamic combat scenarios, our method achieves SOTA performance demonstrating that GROW establishes a new SOTA in both success rate and execution efficiency. Notably, our framework exhibits strong generalization to unseen tasks and fosters sophisticated behavioral skills, such as active target reacquisition and distractor-robust GUI operation , proving its effectiveness in learning reusable interaction strategies rather than merely memorizing trajectories.
\end{itemize}

\section{Related Work}
\subsection{Agents in the Open World}
Open-world tasks provide an important testbed for developing general-purpose agents because they require agents to perceive visual observations, reason about dynamic environments, and execute actions over extended interaction sequences. Minecraft is a representative example due to its high degree of freedom and broad task diversity, spanning embodied control, resource manipulation, and precise graphical user interface interactions. Following the seminal work of VPT~\citep{baker2022video} which leverages large-scale expert demonstrations and explores RL for fine-tuning, subsequent studies have advanced agent learning in several directions. Building upon VPT~\citep{baker2022video}, STEVE-1~\citep{lifshitz2023steve} is trained for text-to-behavior generation in Minecraft, enabling users to use text instructions to control agents for completing short-horizon, open-ended tasks relying on raw pixels and low-level controls.
ROCKET-1~\citep{11093403} introduces visual-temporal context prompting to connect high-level VLM reasoning with low-level policy execution for spatially grounded interaction. ROCKET-3~\citep{cai2025scalable} further improves exploration in unseen environments through RL with cross-view reasoning. More recently, research has increasingly focused on VLM agents. For example, JARVIS-VLA~\citep{li-etal-2025-jarvis} adopts staged training to improve task completion in Minecraft, while similar imitation-based VLM agents have also shown effectiveness in other game environments such as \textit{Genshin Impact}~\citep{tan2025lumineopenrecipebuilding} and Steam games~\citep{magne2026nitrogenopenfoundationmodel}. Despite this progress, efficient RL methods for VLM agents remain underexplored, as most existing approaches still depend heavily on imitation learning. Our work addresses this gap by providing a scalable RL framework for training open-world VLM agents.

\subsection{GRPO for Multi-Turn VLM Agents}

GRPO has been widely explored for RL in tasks requiring many rounds of interaction. AgentGym-RL~\citep{Xi2025AgentGymEA} studies multi-turn RL for large language model agents and improves long-horizon decision making with a curriculum over interaction length. InquireMobile~\citep{DBLP:journals/corr/abs-2508-19679} and ColorAgent~\citep{li2025coloragentbuildingrobustpersonalized} extend this paradigm to settings where the agent must request authorization before taking actions or incorporate human instructions during execution. AGENTRL~\citep{zhang2025agentrl} further investigates system-level scheduling and resource allocation for GRPO-based multi-turn training in order to improve training efficiency. However, although these works substantially broaden the study of GRPO in multi-turn settings, they mainly retain a trajectory-level or dialogue-level formulation, where each optimization sample may contain an increasingly long interaction history. 
When directly applied to open-world tasks, such full-trajectory samples can introduce substantial irrelevant noise and lead to excessive context accumulation as trajectories grow longer. Our work addresses this issue by decomposing trajectories into state-action samples, and further provides a surrogate analysis showing that, although this departs from the identical-prompt grouping assumption of standard GRPO, it still preserves a valid and effective relative policy optimization signal.

\section{Method}
\subsection{Notation}
We formulate the process of executing open-world tasks as a Markov decision process (MDP), denoted by $\mathcal{M}=\langle\mathcal{C}, \mathcal{S}, \mathcal{A}, \mathcal{R}, \gamma\rangle$. Here, $\mathcal{C}$ denotes the task space, which contains a set of heterogeneous tasks $\mathcal{C}=\{c_1, c_2, c_3, \ldots\}$, where each task corresponds to a concrete objective such as \textit{kill zombie} or \textit{mine gold ore}. The state space $\mathcal{S}$ denotes the set of states, where each state corresponds to the current observation or a short history of recent observations together with the task instruction. The action space $\mathcal{A}$ consists of primitive keyboard and
mouse operations, such as \texttt{KEYDOWN} and \texttt{MOUSE\_MOVE}, ensuring applicability to both embodied interactions and graphical user interface (GUI) manipulation.
This action space supports both embodied interaction and GUI manipulation.
We provide more details about the action space in Appendix~\ref{appendix:mc}.
A trajectory $\tau$ is defined as a sequence of state-action pairs, i.e. $\tau=\{ (s_1, a_1), (s_1, a_1), \dots, (s_H, a_H) \}$ where $H$ is the length of $\tau$.
We consider a sparse and verifiable reward setting, where $\mathcal{R}(\tau)=1$ only when successful task completion can be verified, and $\mathcal{R}(\tau)=0$ otherwise. Finally, $\gamma \in (0,1)$ denotes the discount factor.

\subsection{GROW Framework}
Figure~\ref{fig:main_picture} provides an overview of GROW, our proposed RL framework. 
During rollout phase, state-action samples are collected by decomposing the rollout trajectories and then compute relative advantages among the state-action samples belonging to the same rollout groups. 
This design preserves the state-action modeling paradigm commonly used in open-world tasks, while avoiding the need to optimize an entire rollout trajectory as a single full-context sample.

\subsubsection{Decomposition of Rollout Trajectories}
\label{sec:method_rl}
During the rollout phase, $G$ parallel environments are instantiated for each task instruction. In each environment, the VLM agent receives states from the environments and selects the corresponding actions to perform the tasks. Then a group of trajectories $\{\tau_{i}\}_{i=1}^G$ are collected in the $G$ environments, where each trajectory is treated as a single training sample in standard GRPO. For open-world tasks that require many interaction turns, such full-trajectory samples can introduce substantial irrelevant information and lead to excessive context accumulation, ultimately degrading the quality of policy gradient estimation.

To address this issue, we decompose each collected trajectory into a set of fine-grained state-action samples. Specifically, each trajectory $\tau=\{(s_1, a_1), (s_2, a_2), \dots , (s_H, a_H)\}$ is decomposed in a step-by-step manner, where each individual transition serves as an independent optimization unit.
To assign learning signals to the decomposed samples, we propagate the sparse episodic reward backward along each trajectory with a discount factor:
\begin{equation}
    r_{i,t} = \gamma^{H_{i}-t} \mathcal{R}(\tau_{i}),
    \label{equ:discount}
\end{equation}
where $\gamma \in (0,1)$ is the discount factor. This temporal discounting ensures that state-action samples closer to task completion receive stronger learning signals, reflecting their higher causal relevance to the final outcome.
The rollout group is therefore transformed into a group of state-action samples, i.e. $ G_\mathrm{s}=\{(s_{i,t}, a_{i,t}, r_{i,t}) \mid i \in [1,G],\, t \in [1,H_{i}] \}$. By this means, we reformulate the standard GRPO by changing the optimization units from full-trajectory samples to fine-grained state-action samples.

\subsubsection{Policy Optimization with State-Action Samples}
\begin{figure}[t]
  \centering
  \includegraphics[width=\linewidth]{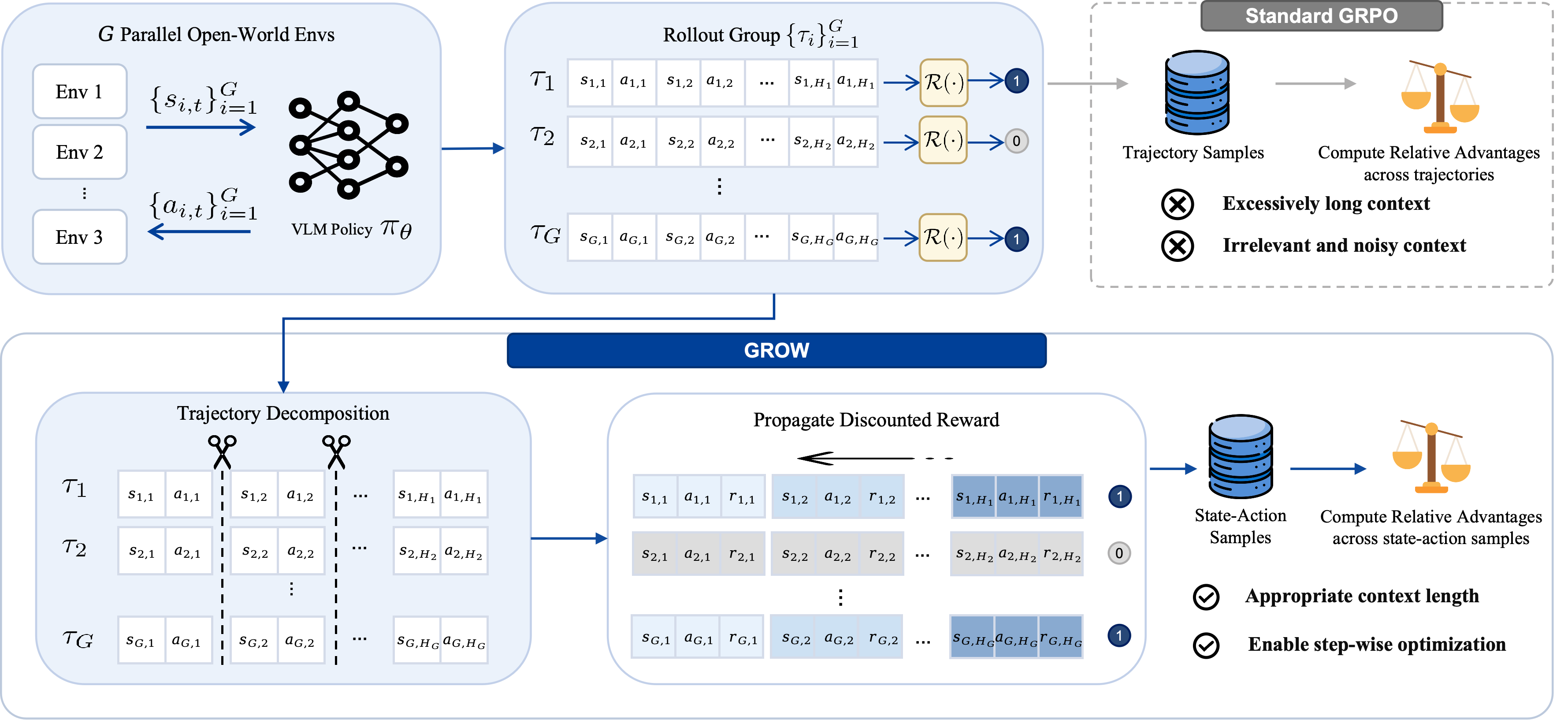}
\caption{
Overview of GROW, a RL framework for open-world VLM agents.  Standard GRPO collects full-trajectory samples from the $G$ rollouts, which often leads to excessively long and noisy contexts in open-world tasks. In the contrast, GROW addresses this issue by decomposing these trajectories into state-action samples, ensuring appropriate context lengths.
}
\label{fig:main_picture}
\end{figure}

Unlike standard GRPO, where relative advantages are computed among the trajectories in the same  groups, GROW computes relative advantages over the decomposed state-action samples within each rollout group. 
As illustrated by the advantage computation module in Figure~\ref{fig:main_picture}, the discounted rewards are normalized across state-action samples derived from the same rollout groups to obtain the advantage:
\begin{equation}
    A_{i,t} = \frac{r_{i,t} - \mu}{\sigma},
\end{equation}
where $\mu$ and $\sigma$ are the mean and standard deviation of the reward set $\{r_{i,t} \mid i \in [1,G],\, t \in [1,H_{i}]\}$. This leads to our training objective, which is defined as:
\begin{equation}
\label{eq:objective}
    \mathcal{J} = \underset{
\substack{
c \sim \mathcal{C} \\
a \sim \pi_{\mathrm{old}}(\cdot \mid s)
}
}{\mathbb{E}} \Biggl\{ \frac{1}{G}\sum_{i=1}^{G} \frac{1}{H_{i}}\sum_{t=1}^{H_{i}}
    \min \left( \rho_{i,t}(\theta) A_{i,t}, \mathrm{clip}(\rho_{i,t}(\theta), 1-\epsilon, 1+\epsilon)A_{i,t} \right) \Biggr\}
\end{equation}
where $\rho_{i,t}(\theta)=\frac{\pi_\theta(a_{i,t} | s_{i,t})}{\pi_{\mathrm{old}}(a_{i,t} | s_{i,t})}$ is the probability ratio.

\subsection{Surrogate Analysis}
\label{subsec:math_gamma_eq_1}

In this section, we provide a surrogate analysis showing that computing relative advantages over decomposed state-action samples can induce a meaningful relative optimization signal, despite departing from the identical-prompt setting of standard GRPO. 

Since the required number of interactions to complete a specific task is generally consistent across 
parallel environments, and we impose a maximum step limit on unsuccessful trajectories, we can approximate the trajectory length for a given task $c$ as a uniform constant $H$, namely $H_i \approx H$. 
We provide the statistics of task step counts in Appendix~\ref{appendix:steps_hist}.
Under this mild assumption, the mean reward over all decomposed state-action samples in the rollout group can be written as
\begin{align}
\mu 
&= \frac{1}{\sum_{i=1}^{G} H_i}\sum_{i=1}^{G}\sum_{t=1}^{H_i} r_{i,t} \nonumber 
= \frac{1}{\sum_{i=1}^{G} H_i}\sum_{i=1}^{G}\sum_{t=1}^{H_i} \gamma^{H_i-t}\mathcal{R}(\tau_i) \nonumber \\
&= \frac{1}{\sum_{i=1}^{G} H_i}\sum_{i=1}^{G}\mathcal{R}(\tau_i)\sum_{t=1}^{H_i}\gamma^{H_i-t} \nonumber \approx \frac{1}{GH}\sum_{i=1}^{G}\mathcal{R}(\tau_i)\sum_{t=1}^{H}\gamma^{H-t}
\end{align}

We further define
\begin{equation}
C_\gamma
=
\frac{1}{H}\sum_{t=1}^{H}\gamma^{H-t}
=
\frac{1-\gamma^{H}}{H(1-\gamma)},
\qquad \gamma \in (0,1)
\end{equation}
which is the average temporal discount coefficient within a trajectory. 
Because $\gamma \in (0,1)$, it immediately follows that
$0 < C_\gamma < 1$.
Let $S$ denotes the average trajectory-level return in the current rollout group, we have $S = \frac{1}{G}\sum_{i=1}^{G}\mathcal{R}(\tau_i)$.
Then the mean reward admits the compact form $\mu \approx C_\gamma S$.

For clarity, we analyze the centered reward $r_{i,t}-\mu$ and omit the standard-deviation term $\sigma$, which acts as a shared positive scaling factor within each rollout group and thus does not affect the  preference.
Under this simplification, the centered reward for each decomposed state-action sample becomes
\begin{equation}
    \hat{A}_{i,t} = r_{i,t} - \mu = \gamma^{H_i-t}\mathcal{R}(\tau_i) - C_\gamma S
    \label{eq:discount_centered_adv}
\end{equation}
 
Substituting Equation~\ref{eq:discount_centered_adv} into the objective in Equation~\ref{eq:objective} (clipping is omitted for brevity) gives
\begin{align}
\label{eq:discounted_split}
\mathcal{J}
&=
\underset{ \substack{ c \sim \mathcal{C} \\ a \sim \pi_{\mathrm{old}}(\cdot \mid s) } }{\mathbb{E}}
\left[
\frac{1}{G}\sum_{i=1}^{G}\frac{1}{H_i}\sum_{t=1}^{H_i}
\rho_{i,t}
\left(
\gamma^{H_i-t}\mathcal{R}(\tau_i)-C_\gamma S
\right)
\right] \nonumber \\
&=
\underset{ \substack{ c \sim \mathcal{C} \\ a \sim \pi_{\mathrm{old}}(\cdot \mid s) } }{\mathbb{E}}
\left[
\frac{1}{G}\sum_{i=1}^{G}\frac{1}{H_i}\sum_{t=1}^{H_i}
C_\gamma \rho_{i,t}
\bigl(\mathcal{R}(\tau_i)-S\bigr)
\right] \nonumber \\
&\quad +
\underset{ \substack{ c \sim \mathcal{C} \\ a \sim \pi_{\mathrm{old}}(\cdot \mid s) } }{\mathbb{E}}
\left[
\frac{1}{G}\sum_{i=1}^{G}\frac{1}{H_i}\sum_{t=1}^{H_i}
\rho_{i,t}
\left(
\gamma^{H_i-t}-C_\gamma
\right)\mathcal{R}(\tau_i)
\right]
\end{align}

For convenience, we define that 
\begin{equation}
\label{eq:obj_traj}
\mathcal{J}_{\mathrm{traj}}
=
\underset{ \substack{ c \sim \mathcal{C} \\ a \sim \pi_{\mathrm{old}}(\cdot \mid s) } }{\mathbb{E}}
\left[
\frac{1}{G}\sum_{i=1}^{G}\frac{1}{H_i}\sum_{t=1}^{H_i}
\rho_{i,t}\bigl(\mathcal{R}(\tau_i)-S\bigr)
\right] 
\end{equation}
\begin{equation}
\mathcal{J}_{\mathrm{step}}
=
\underset{ \substack{ c \sim \mathcal{C} \\ a \sim \pi_{\mathrm{old}}(\cdot \mid s) } }{\mathbb{E}}
\left[
\frac{1}{G}\sum_{i=1}^{G}\frac{1}{H_i}\sum_{t=1}^{H_i}
\rho_{i,t}\left(\gamma^{H_i-t}-C_\gamma\right)\mathcal{R}(\tau_i)
\right]
\end{equation}

Then Equation~\ref{eq:discounted_split} can be rewritten as $
    \mathcal{J}
    =
    C_\gamma \mathcal{J}_{\mathrm{traj}}
    +
    \mathcal{J}_{\mathrm{step}}$.
On the one hand, $\mathcal{J}_{\mathrm{traj}}$ preserves the trajectory-level relative preference because the update signal still depends on the trajectory reward $\mathcal{R}(\tau_i)-S$. 
Therefore, it reinforces successful trajectories at the trajectory level.
On the other hand, $\mathcal{J}_{\mathrm{step}}$ captures temporal discounting effects only on successful trajectories. 
As a result, this term refines credit assignment within successful trajectories by emphasizing decisions that are more directly related to the final success.

Therefore, although the samples in each group are conditioned on different local states rather than an identical prompt context, the objective remains effective: it still compares trajectories through their relative rewards while further refining the update signal at the step level.

\section{Experiments}

\label{sec:experiments}
\subsection{Experimental Setup}

\textbf{Environment} 
We conduct experiments in Minecraft (Java Edition, v1.16.5). In Minecraft, the agent's observation space is strictly limited to first-person view RGB images with a resolution of $360 \times 640 \times 3$, without access to any auxiliary state information. The action space consists of discretized human-like interface commands, including mouse movements, clicks, and keyboard inputs, to simulate authentic human gameplay. 

\textbf{Benchmarks and Task Categorization}
To comprehensively assess the agent's multimodal understanding and fine-grained manipulation capabilities, we use the MCU benchmark~\citep{lin2023mcu}, which includes over 800 tasks. The benchmark comprises distinct task categories, each targeting specific skill sets: (1) \textit{Embodied tasks}: The agent needs to navigate to the position of the target blocks and then use tools to mine or chop blocks. This task focuses on 3D spatial awareness and navigation. Agents must identify target textures within a complex voxel environment and utilize specific tools to excavate target blocks. (2) \textit{GUI tasks}: The agent is requested to create the target items with a furnace or crafting table. This task tests precise 2D grid manipulation and logical reasoning to convert materials. It further introduces temporal constraints, requiring agents to manage wait times during smelting processes. (3) \textit{Combat tasks}: The agent needs to kill the target entities. This task focuses on adversarial strategies in highly dynamic environments. Unlike static tasks, this requires the agent to track and defeat actively hostile moving targets. 

\textbf{Evaluation Metrics} We follow~\citep{wang2025openha,zhou2026main} and employ three primary metrics to assess the performance and efficiency of the agents across different task categories: (1) \textit{Steps}: The average number of interaction steps required to finish a task. (2) \textit{ASR}: The overall average success rate over all tasks in that category, measuring the comprehensive task completion capability of the model. Each task is evaluated over a minimum of 3 independent episodes to ensure reliable performance estimates. To accurately reflect performance variance, the ASR metrics are reported as the mean value alongside their standard deviation. 

\subsection{Implementation Details} 
\label{sec:implementation_details}

Given that most VLM agents in Table~\ref{tab:main_results} are built upon Qwen2-VL-7B-Instruct~\citep{wang2024qwen2vlenhancingvisionlanguagemodels}, we also adopt it as the base model for our agent to ensure a fair comparison. We utilize 3M state-action samples to initialize the VLM agent using LLaMA-Factory framework~\citep{zheng2024llamafactory} with 8 H200 GPUs for about 3 days. We select 8 tasks for RL training to cover all three task categories: 2 embodied tasks (\textit{mine block diorite}, \textit{mine birch log}), 4 GUI tasks (\textit{craft furnace}, \textit{craft iron pickaxe}, \textit{smelt item cooked porkchop}, \textit{smelt gold ingot}) and 2 combat tasks (\textit{kill skeleton}, \textit{kill blaze}). Task success is automatically determined by a built-in environment verifier, which monitors game state transitions such as inventory changes and entity defeat events to produce binary rewards without human intervention. We trained our model in a total different world in Minecraft by selecting world seeds different from those used in evaluation. The RL pipeline is implemented upon the open-source verl framework~\citep{sheng2024hybridflow} to facilitate efficient distributed optimization. We conduct all training procedures on a unified compute node equipped with 8 H200 GPUs. The model undergoes optimization for a total of 240 iterations for about 5 days with a discount factor of $\gamma = 0.995$ to ensure stable and effective convergence across all evaluated tasks. More details about the dataset for policy initialization and hyperparameters are in Appendix~\ref{appendix:dataset} and~\ref{appendix:hyperparameters}. 
\subsection{Main Results}

\begin{table}[]
    \centering
    \caption{
    Evaluation results of Minecraft agents on over 800 tasks.
    ASR is reported as the mean value alongside its standard deviation.
    The best performance is marked in \textbf{bold}, and the second-best performance is \underline{underlined}.
    }
    \label{tab:main_results}
    \resizebox{\linewidth}{!}{
    \renewcommand\arraystretch{1.2}
    \begin{tabular}{@{}llcccccccc@{}}
\toprule
 &  & \multicolumn{2}{c}{Embodied Tasks} &  & \multicolumn{2}{c}{GUI Tasks} &  & \multicolumn{2}{c}{Combat Tasks} \\ 
\cmidrule(lr){3-4} \cmidrule(lr){6-7} \cmidrule(l){9-10} 
\multirow{-2}{*}{Model} & \multirow{-2}{*}{Size} 
& Steps & ASR (All) 
&  & Steps & ASR (All) 
&  & Steps & ASR (All) \\ 
\midrule

VPT~\citep{baker2022video} & 248M 
& 377 & 6.0$^{\pm11.4}$ 
&  & 398 & 0.8$^{\pm3.3}$ 
&  & 396 & 3.6$^{\pm7.7}$ \\

STEVE-1~\citep{lifshitz2023steve} & 248M 
& 384 & 8.0$^{\pm17.0}$ 
&  & 391 & 3.2$^{\pm8.4}$ 
&  & 395 & 3.9$^{\pm12.0}$ \\

ROCKET-1~\citep{cai2025rocket} & 72B 
& 392 & 18.9$^{\pm24.3}$ 
&  & N/A & 0.0 
&  & 320 & 27.9$^{\pm29.3}$ \\

JARVIS-VLA~\citep{li-etal-2025-jarvis} & 7B 
& 305 & 30.0$^{\pm35.4}$ 
&  & 339 & 25.1$^{\pm23.9}$ 
&  & 352 & 18.5$^{\pm22.7}$ \\

UI-TARS-1.5~\citep{shi2025mobileguirladvancingmobilegui} & 7B 
& 290 & 42.1$^{\pm20.4}$ 
&  & 320 & 36.7$^{\pm17.2}$ 
&  & 346 & 31.0$^{\pm16.4}$ \\

OpenHA~\citep{wang2025openha} & 7B 
& 287 & 30.1$^{\pm13.9}$ 
&  & 314 & 32.5$^{\pm9.2}$ 
&  & 316 & 31.9$^{\pm13.7}$ \\

Game-TARS~\citep{wang2025game} & 7B 
& 373 & \underline{50.4$^{\pm20.7}$} 
&  & 406 & \underline{39.1$^{\pm27.5}$} 
&  & 372 & 38.1$^{\pm4.6}$ \\

MAIN-VLA~\citep{zhou2026main} & 7B 
& \underline{263} & 32.8$^{\pm15.4}$ 
&  & \underline{291} & 34.4$^{\pm14.4}$ 
&  & \underline{248} & \underline{39.2$^{\pm16.2}$} \\ 

\rowcolor{rowlightblue}
\textbf{Ours} & \textbf{7B} 
& \textbf{128} & \textbf{59.6$^{\pm37.2}$} 
&  & \textbf{248} & \textbf{68.4$^{\pm29.0}$} 
&  & \textbf{172} & \textbf{49.0$^{\pm35.7}$} \\ 
\bottomrule
\end{tabular}
}
\end{table}

Table~\ref{tab:main_results} shows that GROW achieves the best performance across all three Minecraft task categories.
In embodied tasks, our agent improves the previous best ASR from 50.4\% to 59.6\%, with a gain of 9.2 percentage points.
This improvement suggests that GROW strengthens the agent's ability to navigate toward target objects after detection and to persistently use tools until the task is completed.
In GUI tasks, our RL framework yields the largest gain, improving ASR from 39.1\% to 68.4\%.
This gain mainly comes from improved recognition of target items and better mastery of the key interaction procedures in crafting and smelting.
In combat tasks, our agent improves ASR from 39.2\% to 49.0\%, indicating stronger decision-making in highly dynamic environments where the agent must continuously track enemies, adjust its position, and select appropriate actions under changing visual states.

Besides improving success rates, GROW also substantially reduces the number of steps required to complete tasks.
Compared with the previous best step counts, our agent reduces the average steps from 263 to 128 on embodied tasks, from 291 to 248 on GUI tasks, and from 248 to 172 on combat tasks. 
These results show that GROW not only teaches the agent how to accomplish diverse Minecraft tasks, but also improves execution efficiency, leading to more direct and effective task completion.

\begin{table}[]
    \centering
    \caption{
    Success rates before and after RL on in-domain and out-domain Minecraft tasks.
    Success rates are reported as the mean value alongside their standard deviation.
    }
    \label{tab:rl_before_after_domain}
    \resizebox{\linewidth}{!}{
    \renewcommand\arraystretch{1.2}
    \begin{tabular}{@{}lccccccc@{}}
\toprule
 & \multicolumn{3}{c}{RL-Trained} &  & \multicolumn{3}{c}{RL-Unseen} \\ 
\cmidrule(lr){2-4} \cmidrule(l){6-8}
Model 
& Embodied Tasks & GUI Tasks & Combat Tasks 
& 
& Embodied Tasks & GUI Tasks & Combat Tasks \\ 
\midrule
Initialized Policy 
& 35.0$^{\pm7.1}$ 
& 10.0$^{\pm8.2}$ 
& 45.0$^{\pm21.2}$ 
& 
& 41.9$^{\pm34.7}$ 
& 21.3$^{\pm24.8}$ 
& 19.8$^{\pm24.4}$ \\

\rowcolor{rowlightblue}
\textbf{Ours} 
& \textbf{90.0$^{\pm14.1}$} 
& \textbf{92.5$^{\pm5.0}$} 
& \textbf{85.0$^{\pm21.2}$} 
& 
& \textbf{59.2$^{\pm37.4}$} 
& \textbf{68.2$^{\pm29.1}$} 
& \textbf{47.5$^{\pm35.9}$}\\
\bottomrule
\end{tabular}
}
\end{table}

\subsection{Generalization to Unseen Tasks via GROW}

As shown in Table~\ref{tab:rl_before_after_domain}, GROW substantially improves the initialized policy on both RL-trained and RL-unseen tasks.
On RL-trained tasks, the success rate increases from 35.0\% to 90.0\% on embodied tasks, from 10.0\% to 92.5\% on GUI tasks, and from 45.0\% to 85.0\% on combat tasks. More importantly, these gains are not limited to tasks used during RL, suggesting that the learned policy refinement is not merely task-specific adaptation.
On RL-unseen tasks, GROW also improves the success rate from 41.9\% to 59.2\% on embodied tasks, from 21.3\% to 68.2\% on GUI tasks, and from 19.8\% to 47.5\% on combat tasks.
Since these tasks are excluded from RL training, the consistent improvements provide direct evidence that GROW strengthens reusable interaction skills rather than memorizing training trajectories.
These results indicate that the agent trained with GROW does not overfit to the RL-trained task set. Instead, it acquires transferable capabilities that generalize to unseen Minecraft tasks across different categories.
\subsection{Abalation Study}

\begin{figure}[t]
  \centering
  \includegraphics[width=\linewidth]{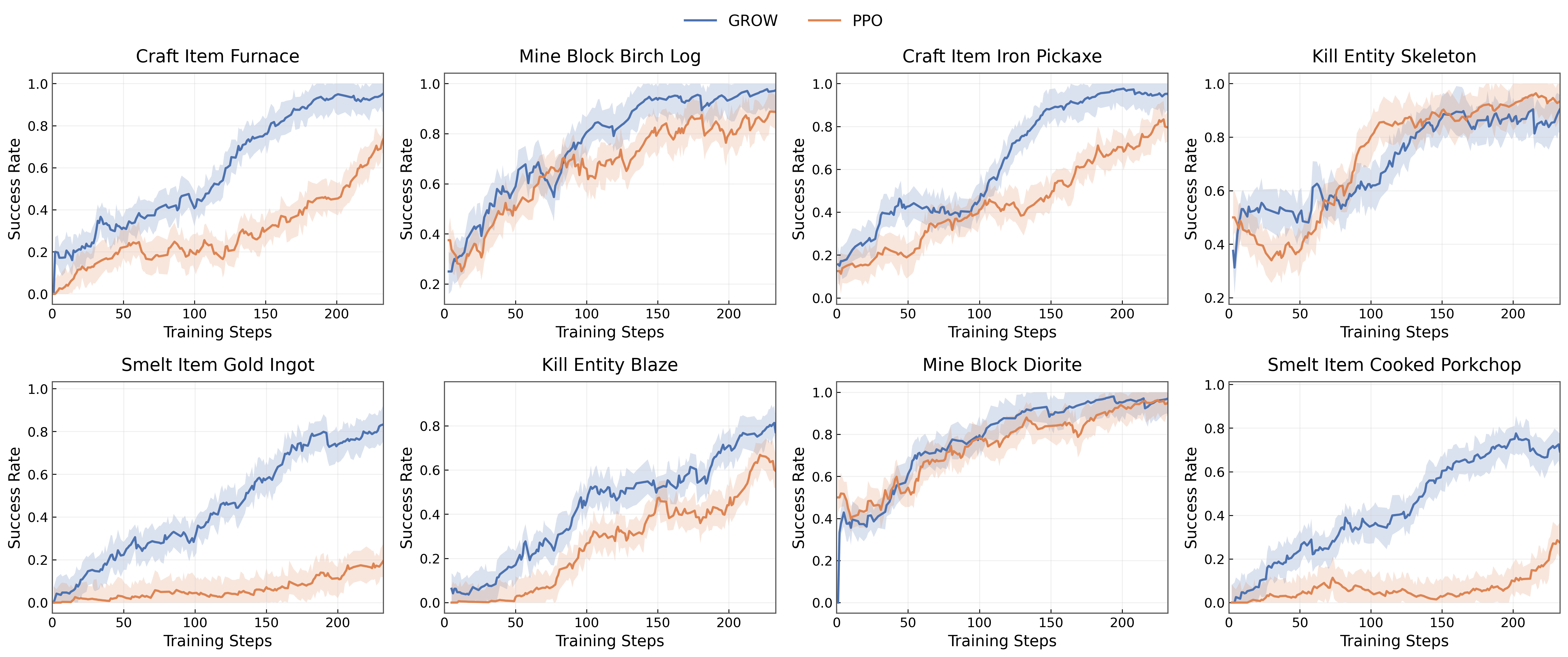}
  
\caption{
Performance comparison between the PPO baseline and GROW, our proposed RL framework, across eight training tasks. The learning curves show that GROW outperforms PPO by achieving significantly higher success rates and faster convergence in most tasks, highlighting the effectiveness of GROW as an RL framework.
}
\label{fig:training_curve}
\end{figure}

\textbf{Comparision with PPO}
Proximal Policy Optimization (PPO) is one of the most widely used RL algorithms and has been extensively adopted before the introduction of GRPO. To examine whether GROW provides stronger policy optimization than this classical RL baseline, we train the initialized policy with PPO and GROW under the same training budget, respectively, and compare their performance in Table~\ref{tab:abalation}. The results show that GROW improves success rates more effectively on RL-trained tasks and also provides stronger generalization to RL-untrained tasks than PPO.

 To further analyze the training dynamics, we compare the success rate curves of GROW and PPO. Figure~\ref{fig:training_curve} reports the success rate curves over the tasks used for RL training. Compared with PPO, GROW improves the success rate faster on most tasks and reaches convergence earlier. These learning curves indicate that GROW shows faster and more stable policy improvement than PPO across most evaluated tasks. 
GROW typically reaches higher success rates earlier in training and maintains a smoother upward trajectory, while PPO often improves more slowly or stays in a low-success plateau for many training steps. 
The advantage is most evident on tasks with delayed procedural rewards and multi-step interaction requirements, where GROW continues to yield measurable gains as training progresses. 
For relatively easier tasks, both methods can eventually approach high success rates, but GROW generally reaches this region earlier. 
\begin{table}[]

    \centering
    \caption{
    Ablation study on RL algorithms and the discount factor $\gamma$. 
    All experiments are trained on the same task set for the same number of training steps to ensure a fair comparison.
    }
    
    \resizebox{\linewidth}{!}{
    \renewcommand\arraystretch{1.2}
    \begin{tabular}{@{}llcccccccc@{}}
\toprule
 &  & \multicolumn{2}{c}{Embodied Tasks} &  & \multicolumn{2}{c}{GUI Tasks} &  & \multicolumn{2}{c}{Combat Tasks} \\ 
\cmidrule(lr){3-4} \cmidrule(lr){6-7} \cmidrule(l){9-10} 
\multirow{-2}{*}{Method} & \multirow{-2}{*}{$\gamma$}
& Steps & ASR (All) 
&  & Steps & ASR (All) 
&  & Steps & ASR (All) \\ 
\midrule

PPO & 0.995
& 177 & $58.2^{\pm30.3}$ 
&  & 580 & $43.9^{\pm24.3}$ 
&  & 181 & $45.4^{\pm32.2}$ \\ 

GROW & 0.9
& 142 & $42.0^{\pm35.2}$ 
&  & 292 & $34.9^{\pm34.4}$ 
&  & 170 & $45.1^{\pm37.8}$ \\ 

GROW & 0.95
& \textbf{122} & 59.0$^{\pm25.9}$ 
&  & 309 & 62.0$^{\pm20.6}$  
&  & \textbf{167} & $46.6^{\pm34.5}$ \\ 

\rowcolor{rowlightblue}
\textbf{GROW} & \textbf{0.995}
& 128 & \textbf{59.6$^{\pm37.2}$} 
&  & \textbf{248} & \textbf{68.4$^{\pm29.0}$} 
&  & 172 & \textbf{49.0$^{\pm35.7}$} \\ 
\bottomrule
\end{tabular}
}
\label{tab:abalation}
\end{table}

\textbf{Abalation on the Discount Factor}
Our RL training framework includes a hyperparameter $\gamma$, which controls how strongly the task-success signal decays when it is propagated backward across interaction steps. Table~\ref{tab:abalation} evaluates how different values of $\gamma$ affect performance, thereby revealing the trade-off between local credit assignment and long-range task completion. For tasks that require relatively few steps, moderately reducing $\gamma$ decreases the value assigned to early steps in overly long trajectories, which suppresses weakly relevant actions and encourages the model to learn more efficient execution strategies. In contrast, for tasks that require many interaction steps, an excessively small $\gamma$ makes the model too short-sighted by weakening useful credit signals from early but necessary actions, and therefore hurts performance.

\subsection{Behavior-Level Analysis}

We conduct a behavior-level analysis on three diagnostic tasks in Minecraft to examine how GROW changes the learned policy. Each task isolates a specific interaction skill and tests whether the state-action formulation of GROW can convert sparse task-level rewards into local policy improvements.

\textbf{Mine Obsidian with Iron Pickaxe}
This task demonstrates stable target fixation. Breaking obsidian with an iron pickaxe requires sustained crosshair alignment over a long mining process, rather than a single correct action.
The baselines cannot complete the task, while GROW reaches 63.3\% success. This suggests that GROW improves precise control across consecutive interaction steps. By decomposing the mining trajectory into state-action samples, GROW reinforces local actions that maintain effective interaction until the block is destroyed.
\begin{wrapfigure}{r}{0.48\textwidth} 
    \centering
    \vspace{-1mm}
    \includegraphics[width=0.48\textwidth]{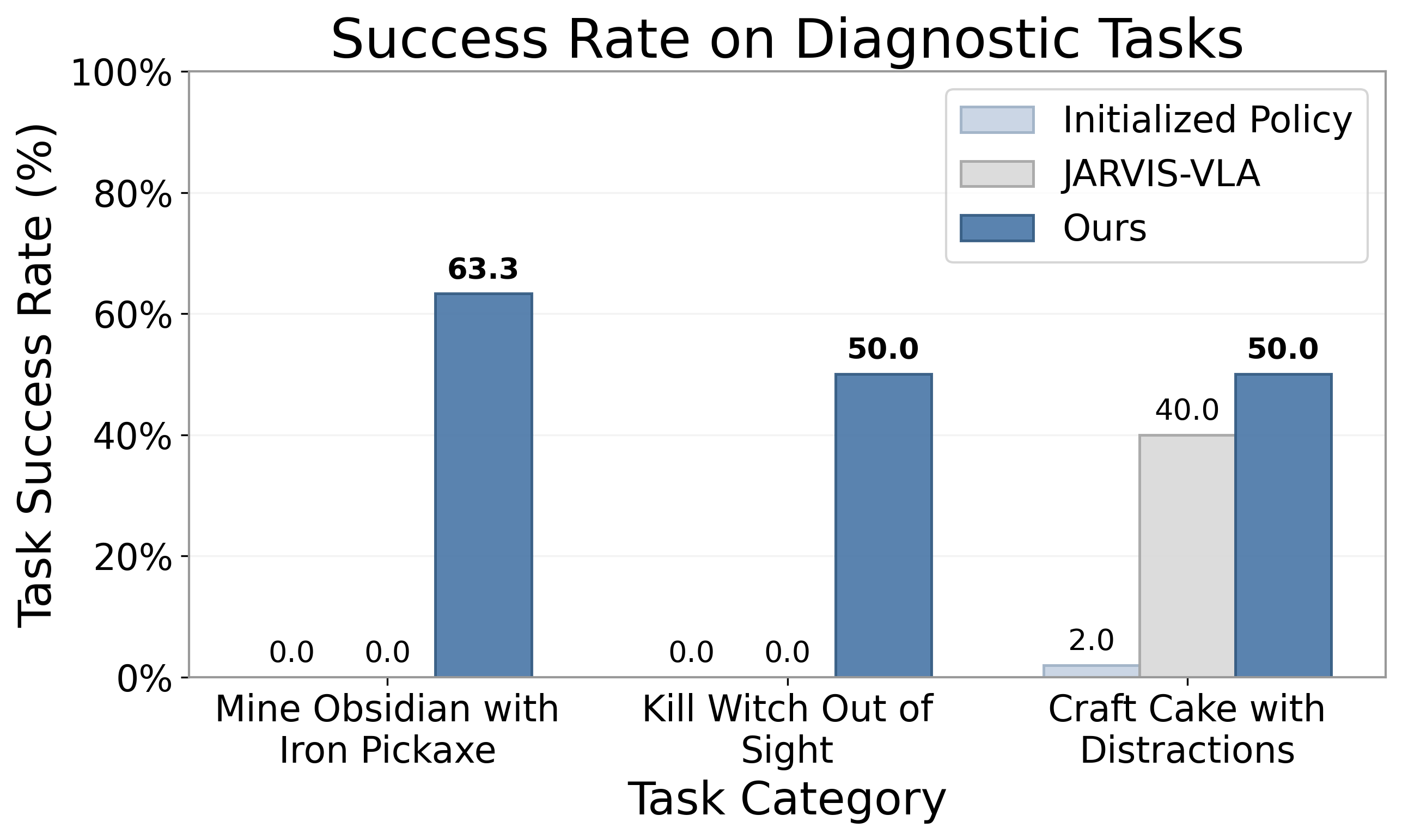}
    \caption{Success rates on diagnostic tasks. GROW improves stable target fixation, active target reacquisition, and distractor-robust GUI operation.}
    \label{fig:diagnostic_task_success}
\end{wrapfigure}

\textbf{Kill Witch Out of Sight}
This task demonstrates active target reacquisition. Since the witch starts outside the field of view, the agent must search, recover visual contact, and continue pursuit as the target moves away. GROW achieves 50.0\% success, whereas both baselines fail completely. This indicates that GROW strengthens not only attack behavior but also camera reorientation, visual tracking, approach, and sustained engagement. We attribute this improvement to the reward assignment induced by trajectory decomposition: trajectories that contain active target-searching behaviors are more likely to achieve final success, so the corresponding action segments receive higher propagated rewards and are repeatedly reinforced during training. As a result, GROW helps the policy acquire the skill of actively searching for and reacquiring targets when visual contact is lost.

\textbf{Craft Cake with Distractions}
This task demonstrates distractor-robust GUI operation. The recipe book and inventory contain many irrelevant items, requiring the agent to select the correct cake recipe under visual clutter. GROW reaches 50.0\% success, compared with 40.0\% for JARVIS-VLA and 2.0\% for the initialized policy. Although JARVIS-VLA captures part of the recipe-book interaction pattern, GROW improves precise recipe selection under distraction. We attribute this improvement to the discounted trajectory reward used in GROW, which encourages the policy to discover more efficient execution paths during training. This signal makes the agent focus more consistently on interface elements that are most relevant to task completion, even when many distractors are present, thereby improving its robustness to visual clutter and increasing the final success rate.

Overall, these diagnostic tasks show that GROW improves stable target fixation, active target reacquisition, and distractor-robust GUI operation. The gains come from optimizing meaningful local decisions within long multi-turn trajectories.

\section{Limitations}
\label{sec:limitations}


Regarding limitations, our current agent mainly performs atomic tasks, lacking a memory module. For long-horizon tasks requiring memory, future work will explore summarizing history after several steps. Additionally, proprietary high-fidelity open-world simulators constrain extending our RL framework to broader training environments. Expanding to diverse accessible environments may further increase the agent's skill repertoire and general applicability.

\section{Conclusion}
\label{sec:conclusion}
We propose GROW, a RL framework for open-world VLM agents. GROW first decomposes trajectories into state-action samples and compute relative advantages between the state-action samples.
By avoiding full-trajectory samples, this design mitigates excessive context accumulation while preserving the relative policy improvement principle of GRPO. 
We also provide surrogate analysis showing that the proposed formulation remains effective when samples in the same optimization group are conditioned on different short-horizon states. 
Experiments on more than 800 Minecraft tasks show that GROW achieves SOTA performance, demonstrating the effectiveness of our RL framework for open-world VLM agents. Across embodied, GUI, and combat tasks, GROW improves both success rate and execution efficiency, while also showing generalization to unseen tasks and fostering reusable interaction skills. 

\bibliographystyle{plainnat}
\bibliography{neurips_2026}

@inproceedings{Xi2025AgentGymEA,
  title={AgentGym: Evaluating and Training Large Language Model-based Agents across Diverse Environments},
  author={Zhiheng Xi and Yiwen Ding and Wenxiang Chen and Boyang Hong and Honglin Guo and Junzhe Wang and Xin Guo and Dingwen Yang and Chenyang Liao and Wei He and Songyang Gao and Luyao Chen and Rui Zheng and Yicheng Zou and Tao Gui and Qi Zhang and Xipeng Qiu and Xuanjing Huang and Zuxuan Wu and Yu-Gang Jiang},
  booktitle={Annual Meeting of the Association for Computational Linguistics},
  year={2025},
  url={https://api.semanticscholar.org/CorpusID:280017701}
}

@article{baker2022video,
  title={Video pretraining (vpt): Learning to act by watching unlabeled online videos},
  author={Baker, Bowen and Akkaya, Ilge and Zhokov, Peter and Huizinga, Joost and Tang, Jie and Ecoffet, Adrien and Houghton, Brandon and Sampedro, Raul and Clune, Jeff},
  journal={Advances in Neural Information Processing Systems},
  volume={35},
  pages={24639--24654},
  year={2022}
}

@inproceedings{li-etal-2025-jarvis,
    title = "{JARVIS}-{VLA}: Post-Training Large-Scale Vision Language Models to Play Visual Games with Keyboards and Mouse",
    author = "Li, Muyao  and
      Wang, Zihao  and
      He, Kaichen  and
      Ma, Xiaojian  and
      Liang, Yitao",
    editor = "Che, Wanxiang  and
      Nabende, Joyce  and
      Shutova, Ekaterina  and
      Pilehvar, Mohammad Taher",
    booktitle = "Findings of the Association for Computational Linguistics: ACL 2025",
    month = jul,
    year = "2025",
    address = "Vienna, Austria",
    publisher = "Association for Computational Linguistics",
    url = "https://aclanthology.org/2025.findings-acl.920/",
    doi = "10.18653/v1/2025.findings-acl.920",
    pages = "17878--17899",
    ISBN = "979-8-89176-256-5"
}

@article{shao2024deepseekmath,
  title={Deepseekmath: Pushing the limits of mathematical reasoning in open language models},
  author={Shao, Zhihong and Wang, Peiyi and Zhu, Qihao and Xu, Runxin and Song, Junxiao and Bi, Xiao and Zhang, Haowei and Zhang, Mingchuan and Li, YK and Wu, Yang and others},
  journal={arXiv preprint arXiv:2402.03300},
  year={2024}
}

@article{lifshitz2023steve,
  title={Steve-1: A generative model for text-to-behavior in minecraft},
  author={Lifshitz, Shalev and Paster, Keiran and Chan, Harris and Ba, Jimmy and McIlraith, Sheila},
  journal={Advances in Neural Information Processing Systems},
  volume={36},
  pages={69900--69929},
  year={2023}
}

@article{cai2025scalable,
  title={Scalable Multi-Task Reinforcement Learning for Generalizable Spatial Intelligence in Visuomotor Agents},
  author={Cai, Shaofei and Mu, Zhancun and Xia, Haiwen and Zhang, Bowei and Liu, Anji and Liang, Yitao},
  journal={arXiv preprint arXiv:2507.23698},
  year={2025}
}

@misc{tan2025lumineopenrecipebuilding,
      title={Lumine: An Open Recipe for Building Generalist Agents in 3D Open Worlds}, 
      author={Weihao Tan and Xiangyang Li and Yunhao Fang and Heyuan Yao and Shi Yan and Hao Luo and Tenglong Ao and Huihui Li and Hongbin Ren and Bairen Yi and Yujia Qin and Bo An and Libin Liu and Guang Shi},
      year={2025},
      eprint={2511.08892},
      archivePrefix={arXiv},
      primaryClass={cs.AI},
      url={https://arxiv.org/abs/2511.08892}, 
}

@misc{magne2026nitrogenopenfoundationmodel,
      title={NitroGen: An Open Foundation Model for Generalist Gaming Agents}, 
      author={Loïc Magne and Anas Awadalla and Guanzhi Wang and Yinzhen Xu and Joshua Belofsky and Fengyuan Hu and Joohwan Kim and Ludwig Schmidt and Georgia Gkioxari and Jan Kautz and Yisong Yue and Yejin Choi and Yuke Zhu and Linxi "Jim" Fan},
      year={2026},
      eprint={2601.02427},
      archivePrefix={arXiv},
      primaryClass={cs.CV},
      url={https://arxiv.org/abs/2601.02427}, 
}

@article{zhang2025agentrl,
  title={AgentRL: Scaling Agentic Reinforcement Learning with a Multi-Turn, Multi-Task Framework},
  author={Zhang, Hanchen and Liu, Xiao and Lv, Bowen and Sun, Xueqiao and Jing, Bohao and Iong, Iat Long and Hou, Zhenyu and Qi, Zehan and Lai, Hanyu and Xu, Yifan and others},
  journal={arXiv preprint arXiv:2510.04206},
  year={2025}
}

@article{wang2025openha,
  title={Openha: A series of open-source hierarchical agentic models in minecraft},
  author={Wang, Zihao and Li, Muyao and He, Kaichen and Wang, Xiangyu and Mu, Zhancun and Liu, Anji and Liang, Yitao},
  journal={arXiv preprint arXiv:2509.13347},
  year={2025}
}

@article{lin2023mcu,
  title={Mcu: A task-centric framework for open-ended agent evaluation in minecraft},
  author={Lin, Haowei and Wang, Zihao and Ma, Jianzhu and Liang, Yitao},
  journal={arXiv preprint arXiv:2310.08367},
  year={2023}
}

@misc{bai2025qwen25vltechnicalreport,
      title={Qwen2.5-VL Technical Report}, 
      author={Shuai Bai and Keqin Chen and Xuejing Liu and Jialin Wang and Wenbin Ge and Sibo Song and Kai Dang and Peng Wang and Shijie Wang and Jun Tang and Humen Zhong and Yuanzhi Zhu and Mingkun Yang and Zhaohai Li and Jianqiang Wan and Pengfei Wang and Wei Ding and Zheren Fu and Yiheng Xu and Jiabo Ye and Xi Zhang and Tianbao Xie and Zesen Cheng and Hang Zhang and Zhibo Yang and Haiyang Xu and Junyang Lin},
      year={2025},
      eprint={2502.13923},
      archivePrefix={arXiv},
      primaryClass={cs.CV},
      url={https://arxiv.org/abs/2502.13923}, 
}

@inproceedings{zheng2024llamafactory,
  title={LlamaFactory: Unified Efficient Fine-Tuning of 100+ Language Models},
  author={Yaowei Zheng and Richong Zhang and Junhao Zhang and Yanhan Ye and Zheyan Luo and Zhangchi Feng and Yongqiang Ma},
  booktitle={Proceedings of the 62nd Annual Meeting of the Association for Computational Linguistics (Volume 3: System Demonstrations)},
  address={Bangkok, Thailand},
  publisher={Association for Computational Linguistics},
  year={2024},
  url={http://arxiv.org/abs/2403.13372}
}

@article{wang2025game,
  title={Game-tars: Pretrained foundation models for scalable generalist multimodal game agents},
  author={Wang, Zihao and Li, Xujing and Ye, Yining and Fang, Junjie and Wang, Haoming and Liu, Longxiang and Liang, Shihao and Lu, Junting and Wu, Zhiyong and Feng, Jiazhan and others},
  journal={arXiv preprint arXiv:2510.23691},
  year={2025}
}

@article{luo2025navimaster,
  title={Navimaster: Learning a unified policy for gui and embodied navigation tasks},
  author={Luo, Zhihao and Yan, Wentao and Gong, Jingyu and Wang, Min and Zhang, Zhizhong and Wang, Xuhong and Xie, Yuan and Tan, Xin},
  journal={arXiv preprint arXiv:2508.02046},
  year={2025}
}

@article{DBLP:journals/corr/abs-2508-19679,
  publtype={informal},
  author={Qihang Ai and Pi Bu and Yue Cao and Yingyao Wang and Jihao Gu and Jingxuan Xing and Zekun Zhu and Wei Jiang and Zhicheng Zheng and Jun Song and Yuning Jiang and Bo Zheng},
  title={InquireMobile: Teaching VLM-based Mobile Agent to Request Human Assistance via Reinforcement Fine-Tuning},
  year={2025},
  month={August},
  cdate={1754006400000},
  journal={CoRR},
  volume={abs/2508.19679},
  url={https://doi.org/10.48550/arXiv.2508.19679}
}

@misc{li2025coloragentbuildingrobustpersonalized,
      title={ColorAgent: Building A Robust, Personalized, and Interactive OS Agent}, 
      author={Ning Li and Qiqiang Lin and Zheng Wu and Xiaoyun Mo and Weiming Zhang and Yin Zhao and Xiangmou Qu and Jiamu Zhou and Jun Wang and Congmin Zheng and Yuanyi Song and Hongjiang Chen and Heyuan Huang and Jihong Wang and Jiaxin Yin and Jingwei Yu and Junwei Liao and Qiuying Peng and Xingyu Lou and Jun Wang and Weiwen Liu and Zhuosheng Zhang and Weinan Zhang},
      year={2025},
      eprint={2510.19386},
      archivePrefix={arXiv},
      primaryClass={cs.MA},
      url={https://arxiv.org/abs/2510.19386}, 
}

@article{sheng2024hybridflow,
  title   = {HybridFlow: A Flexible and Efficient RLHF Framework},
  author  = {Guangming Sheng and Chi Zhang and Zilingfeng Ye and Xibin Wu and Wang Zhang and Ru Zhang and Yanghua Peng and Haibin Lin and Chuan Wu},
  year    = {2024},
  journal = {arXiv preprint arXiv: 2409.19256}
}

@inproceedings{cai2025rocket,
  title={Rocket-1: Mastering open-world interaction with visual-temporal context prompting},
  author={Cai, Shaofei and Wang, Zihao and Lian, Kewei and Mu, Zhancun and Ma, Xiaojian and Liu, Anji and Liang, Yitao},
  booktitle={Proceedings of the Computer Vision and Pattern Recognition Conference},
  pages={12122--12131},
  year={2025}
}

@article{zhou2026main,
  title={MAIN-VLA: Modeling Abstraction of Intention and eNvironment for Vision-Language-Action Models},
  author={Zhou, Zheyuan and Du, Liang and Sun, Zixun and Zhou, Xiaoyu and Ye, Ruimin and Chen, Qihao and Chen, Yinda and Qiu, Lemiao},
  journal={arXiv preprint arXiv:2602.02212},
  year={2026}
}

@misc{Ouyang2026GameWorldTS,
  title={GameWorld: Towards Standardized and Verifiable Evaluation of Multimodal Game Agents},
  author={Mingyu Ouyang and Siyuan Hu and Kevin Qinghong Lin and Hwee Tou Ng and Mike Zheng Shou},
  year={2026},
  url={https://api.semanticscholar.org/CorpusID:287255688}
}

@misc{lynch2022interactivelanguagetalkingrobots,
      title={Interactive Language: Talking to Robots in Real Time}, 
      author={Corey Lynch and Ayzaan Wahid and Jonathan Tompson and Tianli Ding and James Betker and Robert Baruch and Travis Armstrong and Pete Florence},
      year={2022},
      eprint={2210.06407},
      archivePrefix={arXiv},
      primaryClass={cs.RO},
      url={https://arxiv.org/abs/2210.06407}, 
}

@INPROCEEDINGS{11093403,
  author={Cai, Shaofei and Wang, Zihao and Lian, Kewei and Mu, Zhancun and Ma, Xiaojian and Liu, Anji and Liang, Yitao},
  booktitle={2025 IEEE/CVF Conference on Computer Vision and Pattern Recognition (CVPR)}, 
  title={ROCKET-1: Mastering Open-World Interaction with Visual-Temporal Context Prompting}, 
  year={2025},
  volume={},
  number={},
  pages={12122-12131},
  doi={10.1109/CVPR52734.2025.01132}}

@inproceedings{
li2026compassnav,
title={CompassNav: Steering From Path Imitation to Decision Understanding In Navigation},
author={LinFeng Li and Jian Zhao and Yuan Xie and Xin Tan and Xuelong Li},
booktitle={The Fourteenth International Conference on Learning Representations},
year={2026},
url={https://openreview.net/forum?id=eqcDckWHik}
}

@misc{wang2024qwen2vlenhancingvisionlanguagemodels,
      title={Qwen2-VL: Enhancing Vision-Language Model's Perception of the World at Any Resolution}, 
      author={Peng Wang and Shuai Bai and Sinan Tan and Shijie Wang and Zhihao Fan and Jinze Bai and Keqin Chen and Xuejing Liu and Jialin Wang and Wenbin Ge and Yang Fan and Kai Dang and Mengfei Du and Xuancheng Ren and Rui Men and Dayiheng Liu and Chang Zhou and Jingren Zhou and Junyang Lin},
      year={2024},
      eprint={2409.12191},
      archivePrefix={arXiv},
      primaryClass={cs.CV},
      url={https://arxiv.org/abs/2409.12191}, 
}

@misc{hu2026longnavr1horizonadaptivemultiturnrl,
      title={LongNav-R1: Horizon-Adaptive Multi-Turn RL for Long-Horizon VLA Navigation}, 
      author={Yue Hu and Avery Xi and Qixin Xiao and Seth Isaacson and Henry X. Liu and Ram Vasudevan and Maani Ghaffari},
      year={2026},
      eprint={2602.12351},
      archivePrefix={arXiv},
      primaryClass={cs.RO},
      url={https://arxiv.org/abs/2602.12351}, 
}

@misc{zhang2025activevlnactiveexplorationmultiturn,
      title={ActiveVLN: Towards Active Exploration via Multi-Turn RL in Vision-and-Language Navigation}, 
      author={Zekai Zhang and Weiye Zhu and Hewei Pan and Xiangchen Wang and Rongtao Xu and Xing Sun and Feng Zheng},
      year={2025},
      eprint={2509.12618},
      archivePrefix={arXiv},
      primaryClass={cs.RO},
      url={https://arxiv.org/abs/2509.12618}, 
}

@misc{ye2025etpr1evolvingtopologicalplanning,
      title={ETP-R1: Evolving Topological Planning with Reinforcement Fine-tuning for Vision-Language Navigation in Continuous Environments}, 
      author={Shuhao Ye and Sitong Mao and Yuxiang Cui and Xuan Yu and Shichao Zhai and Wen Chen and Shunbo Zhou and Rong Xiong and Yue Wang},
      year={2025},
      eprint={2512.20940},
      archivePrefix={arXiv},
      primaryClass={cs.RO},
      url={https://arxiv.org/abs/2512.20940}, 
}

@misc{shi2025mobileguirladvancingmobilegui,
      title={MobileGUI-RL: Advancing Mobile GUI Agent through Reinforcement Learning in Online Environment}, 
      author={Yucheng Shi and Wenhao Yu and Zaitang Li and Yonglin Wang and Hongming Zhang and Ninghao Liu and Haitao Mi and Dong Yu},
      year={2025},
      eprint={2507.05720},
      archivePrefix={arXiv},
      primaryClass={cs.LG},
      url={https://arxiv.org/abs/2507.05720}, 
}






\newpage
\appendix

\section{Environment and Action Space}
\textbf{Observation Space.} Our model operates under a strict pixel-only constraint, perceiving the environment solely through raw $640 \times 360$ RGB frames without access to any internal game states, coordinates, or metadata. Following the protocol established in JARVIS-VLA~\citep{li-etal-2025-jarvis}, these observations are intended to mirror the authentic experience of a human player; as such, we do not remove or mask any standard on-screen overlays. The visual input includes all native HUD elements—such as the hotbar, health and hunger indicators, and the dynamic hand-swing animations triggered by interactions—forcing the agent to interpret the environment and its own status purely from raw sensory data. To maintain this human-centric perspective, we employ a standard $70^{\circ}$ Field of View and a GUI scale of 2, ensuring the visual distribution remains consistent with typical gameplay.

\textbf{Action Space.} The action space is designed to be both atomic and expressive, mapping directly to the fundamental keyboard and mouse inputs available to a human. Rather than employing high-level macro-actions or simplified APIs, we decompose player behavior into granular primitives. This includes binary operations for movement and interaction (e.g., sprinting, jumping, and attacking) alongside continuous controls for camera orientation (pitch and yaw). By relying on these basic building blocks, the agent is required to learn the composition of complex, multi-step strategies from the ground up. The full set of these operations is summarized in Table \ref{tab:action_space}. In inference, we request agents to predict an action chunk with 4 interaction steps according to their current observation.

\begin{table}[htbp] 
  \caption{Mapping binary primitives to the standard Minecraft controls.}  
  \label{tab:action_space}
  \centering
  \begin{tabular}{ll p{8cm}} 
    \toprule
    \textbf{Action} & \textbf{Human action} & \textbf{Description} \\   
    \midrule
    forward   & W key          & Move forward. \\
    back      & S key          & Move backward. \\
    left      & A key          & Strafe left. \\
    right     & D key          & Strafe right. \\
    jump      & Space key      & Jump. \\
    sneak     & left Shift key & Switch to a slow walking mode. \\
    sprint    & left Ctrl key  & Switch to rapid walking mode. \\
    attack    & left Button    & Destroy blocks (hold down); Attack entity (click once). \\
    use       & right Button   & Place the item currently held or use the block the player is looking at. \\
    drop      & Q key          & Drop a single item from the stack of items the player is currently holding. \\
    hotbar.[1-9] & keys 1-9    & Switch active item to the one in a given hotbar cell. \\
    inventory & E key          & Open/Close the inventory. \\
    yaw       & move Mouse X   & camera movement. \\
    pitch     & move Mouse Y   & camera movement. \\
    \bottomrule
  \end{tabular}   
\end{table}

\label{appendix:mc}


\section{Statistic Evidences}
\label{appendix:steps_hist}
\begin{figure*}[t]
    \centering
    \captionsetup[subfigure]{justification=centering,skip=1pt}
    \begin{subfigure}[t]{0.24\textwidth}
        \centering
        \includegraphics[width=\linewidth,trim=70 42 70 34,clip]{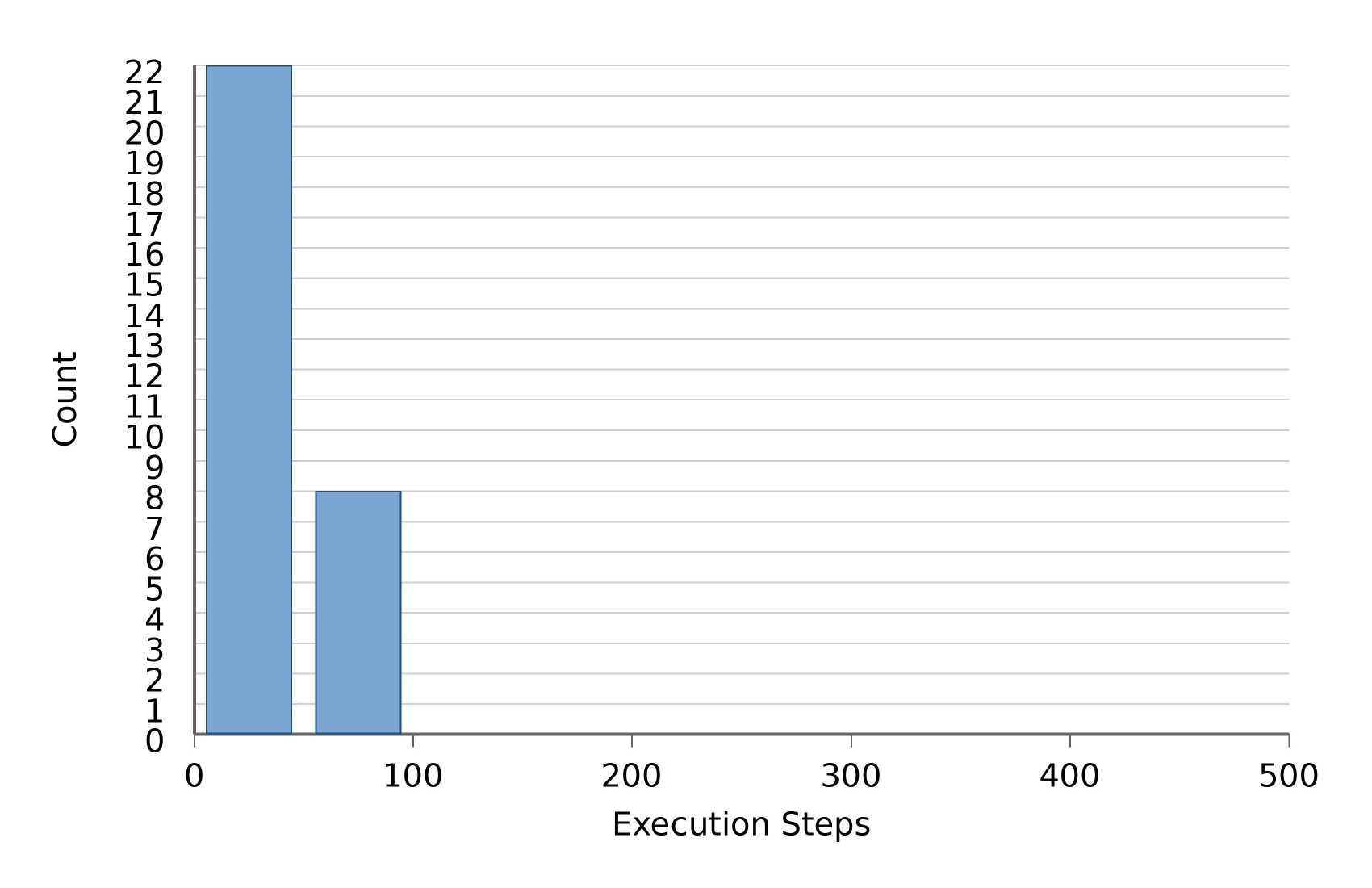}
        \caption{Mine Diorite}
    \end{subfigure}\hfill
    \begin{subfigure}[t]{0.24\textwidth}
        \centering
        \includegraphics[width=\linewidth,trim=70 42 70 34,clip]{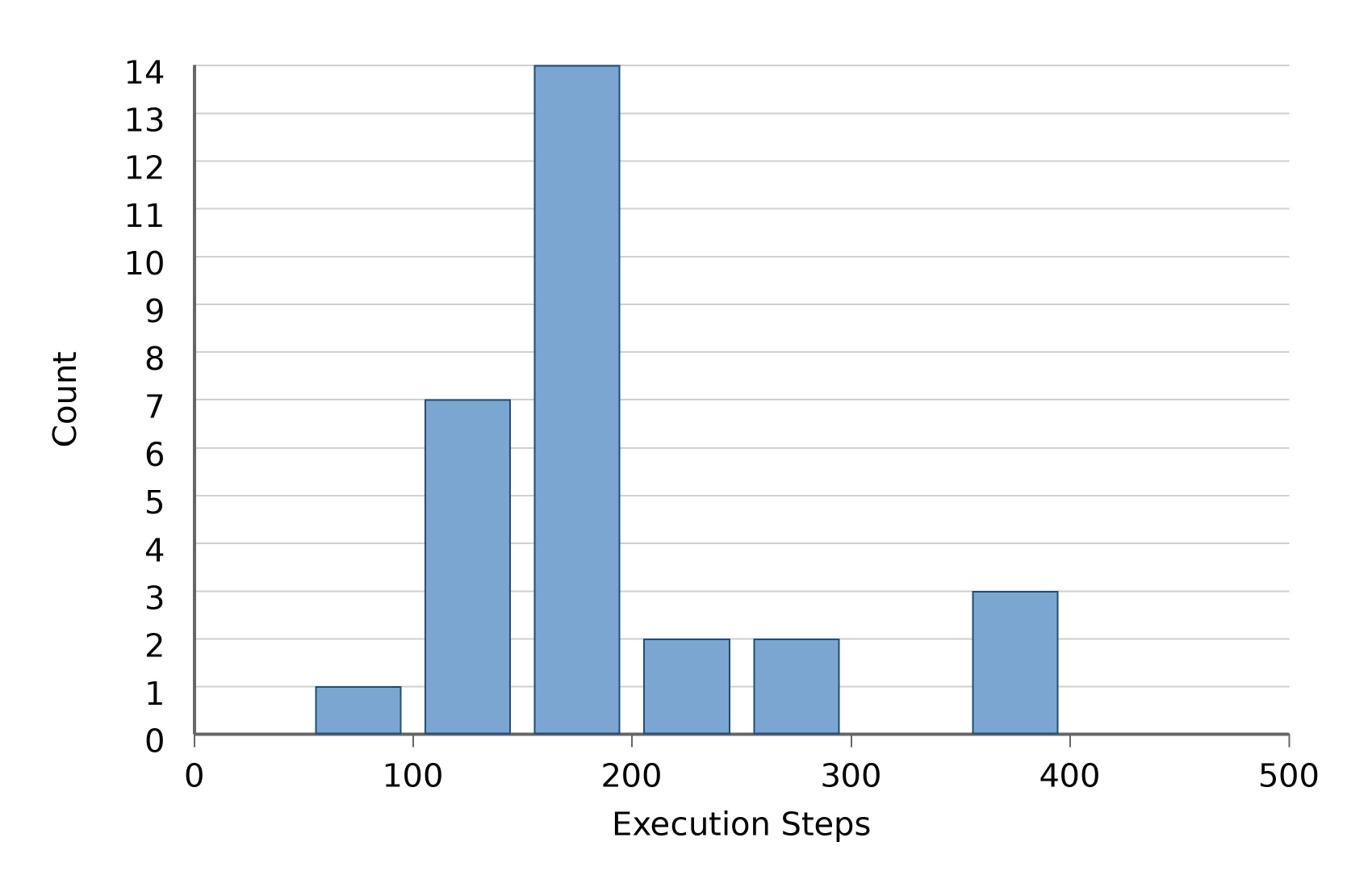}
        \caption{Mine Birch Log}
    \end{subfigure}\hfill
    \begin{subfigure}[t]{0.24\textwidth}
        \centering
        \includegraphics[width=\linewidth,trim=70 42 70 34,clip]{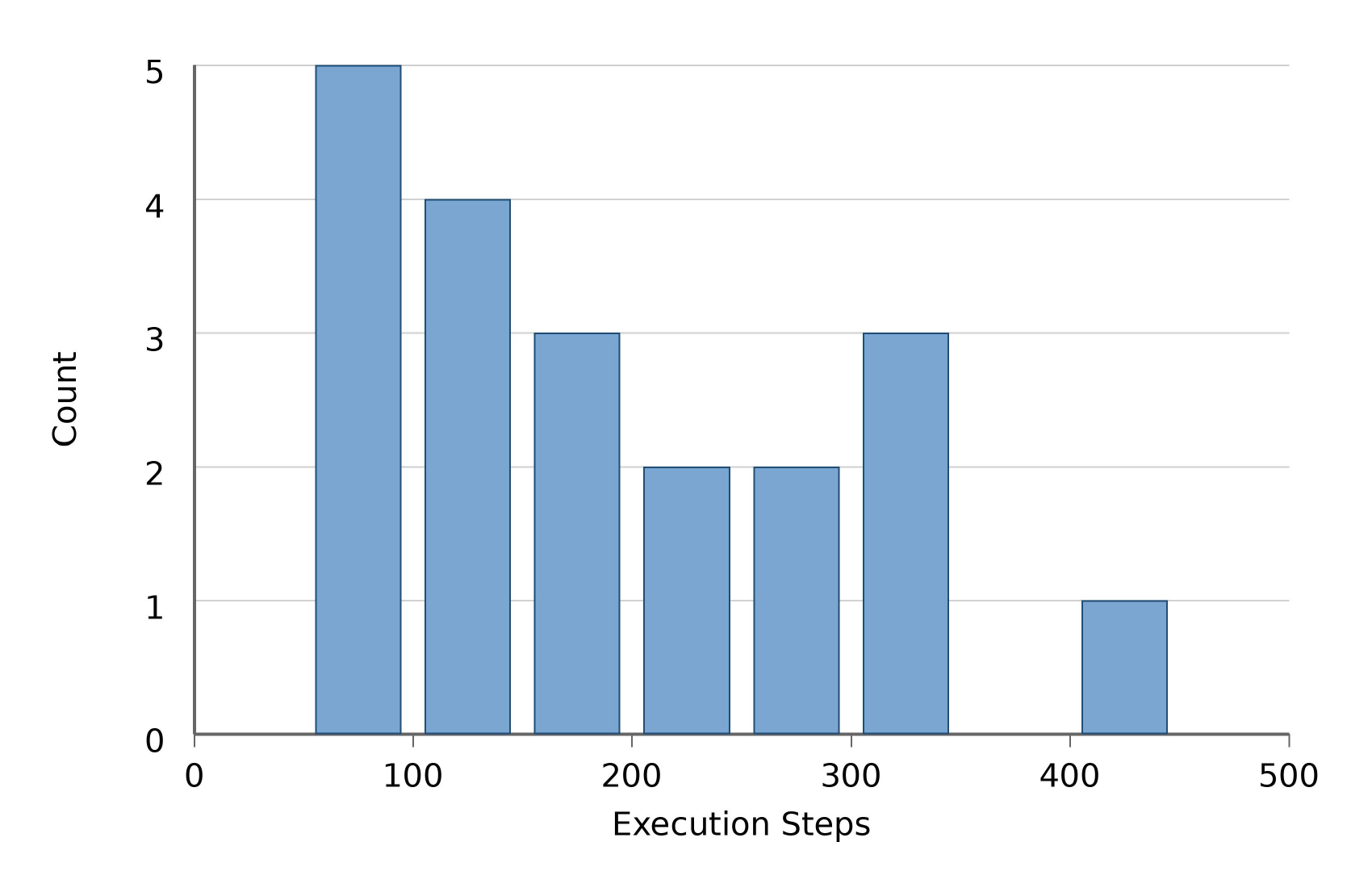}
        \caption{Craft Furnace}
    \end{subfigure}\hfill
    \begin{subfigure}[t]{0.24\textwidth}
        \centering
        \includegraphics[width=\linewidth,trim=70 42 70 34,clip]{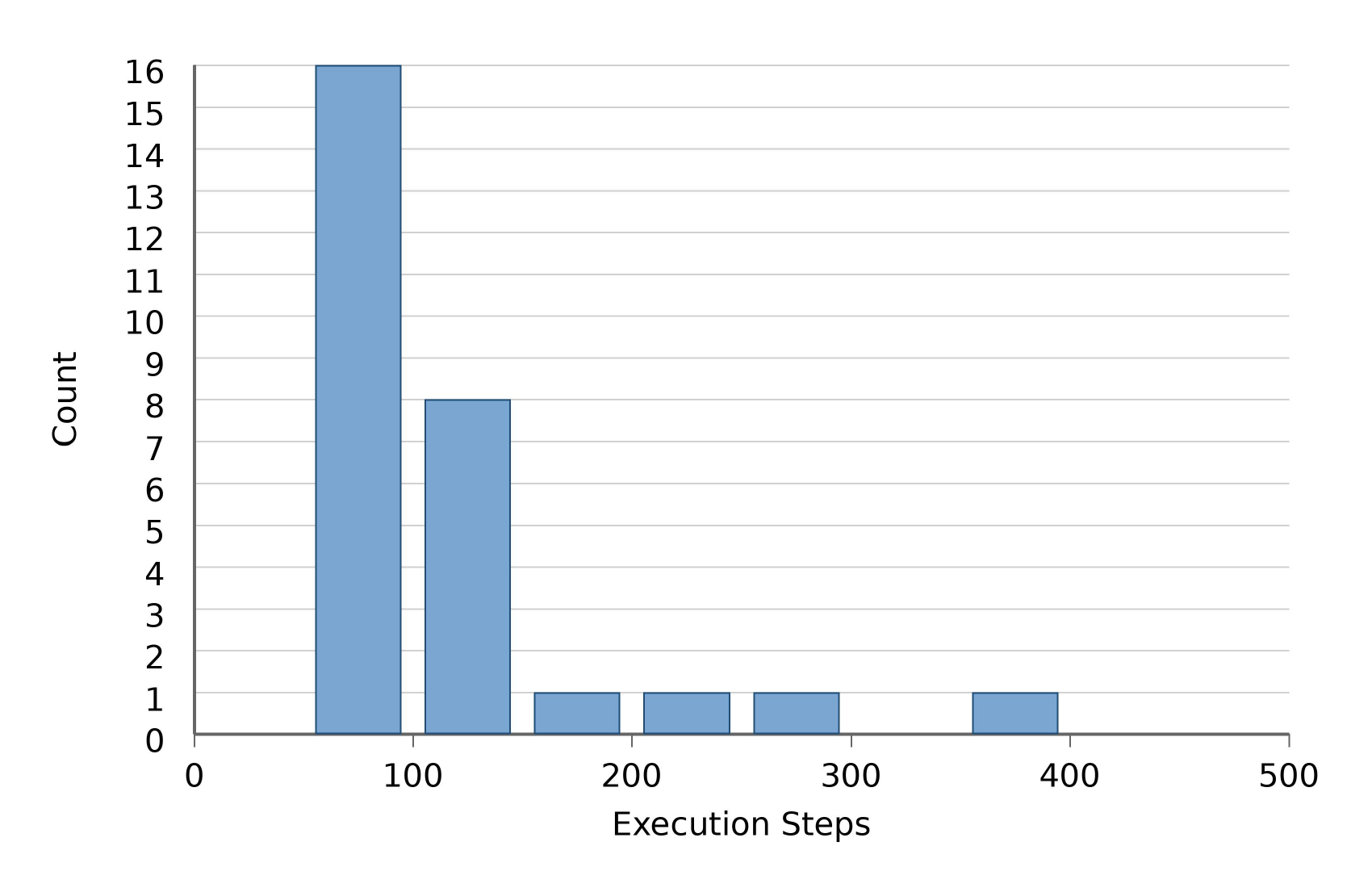}
        \caption{Craft Iron Pickaxe}
    \end{subfigure}

    \vspace{0.25em}

    \begin{subfigure}[t]{0.24\textwidth}
        \centering
        \includegraphics[width=\linewidth,trim=70 42 70 34,clip]{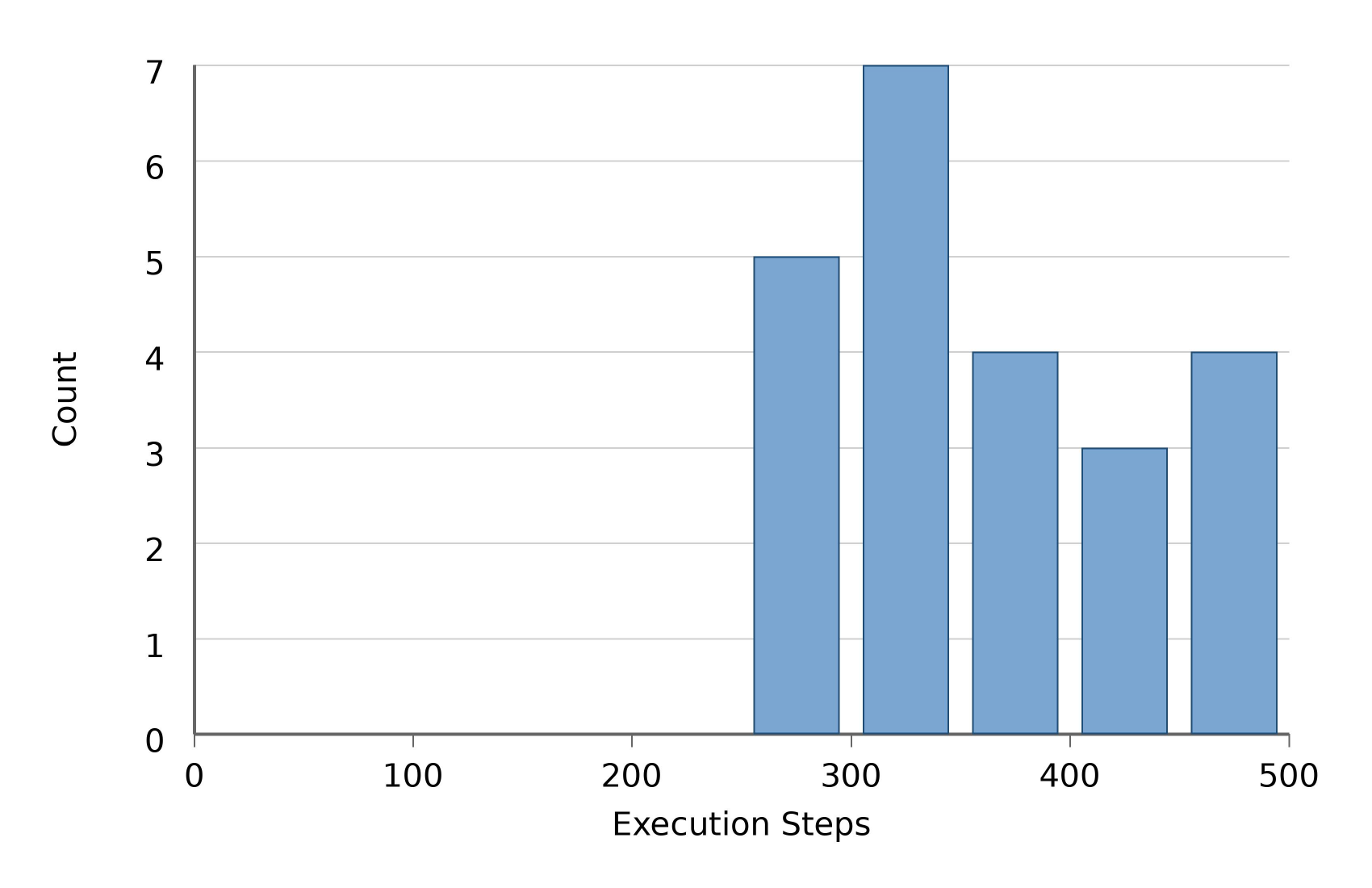}
        \caption{Smelt Cooked Porkchop}
    \end{subfigure}\hfill
    \begin{subfigure}[t]{0.24\textwidth}
        \centering
        \includegraphics[width=\linewidth,trim=70 42 70 34,clip]{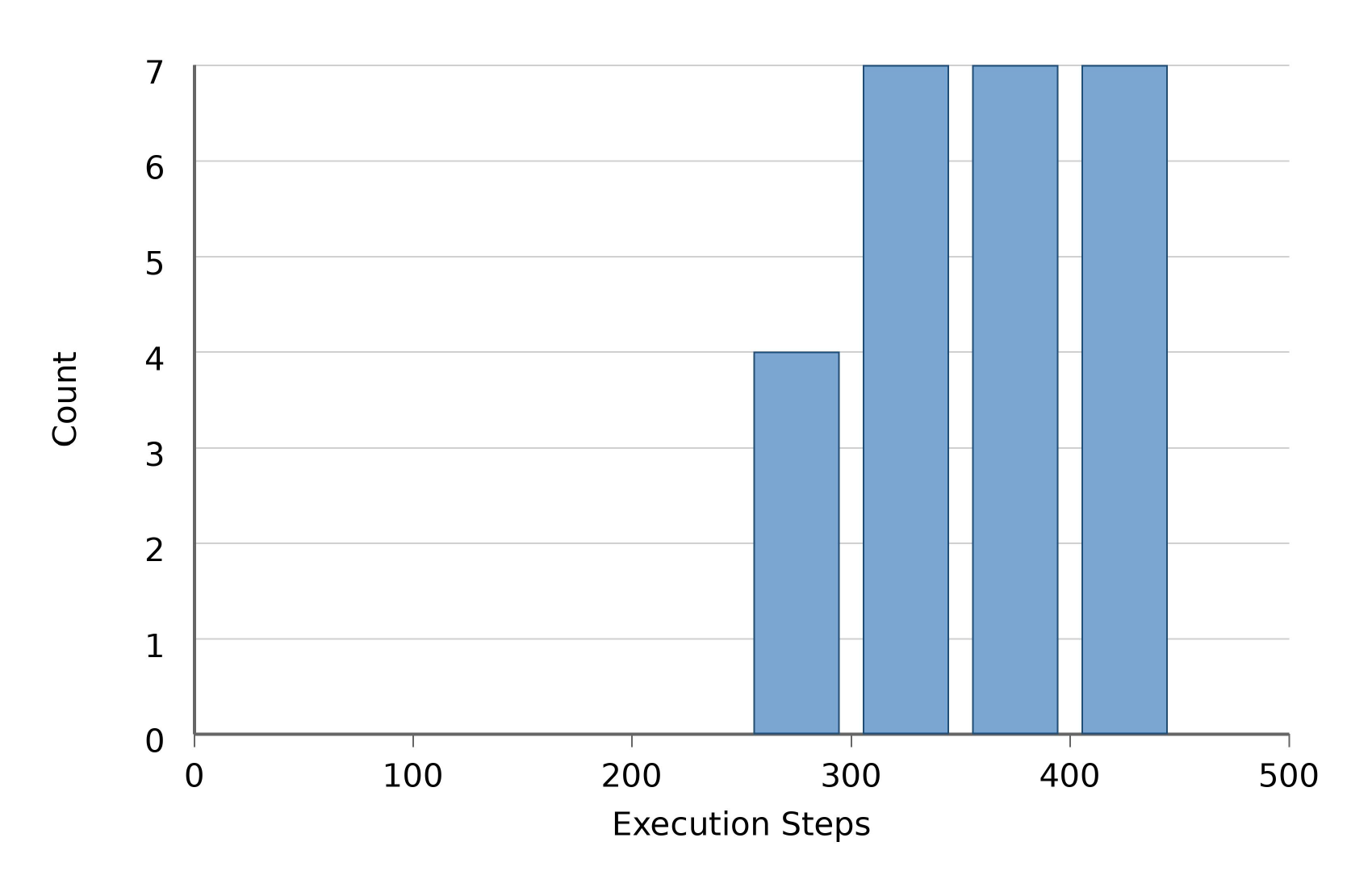}
        \caption{Smelt Gold Ingot}
    \end{subfigure}\hfill
    \begin{subfigure}[t]{0.24\textwidth}
        \centering
        \includegraphics[width=\linewidth,trim=70 42 70 34,clip]{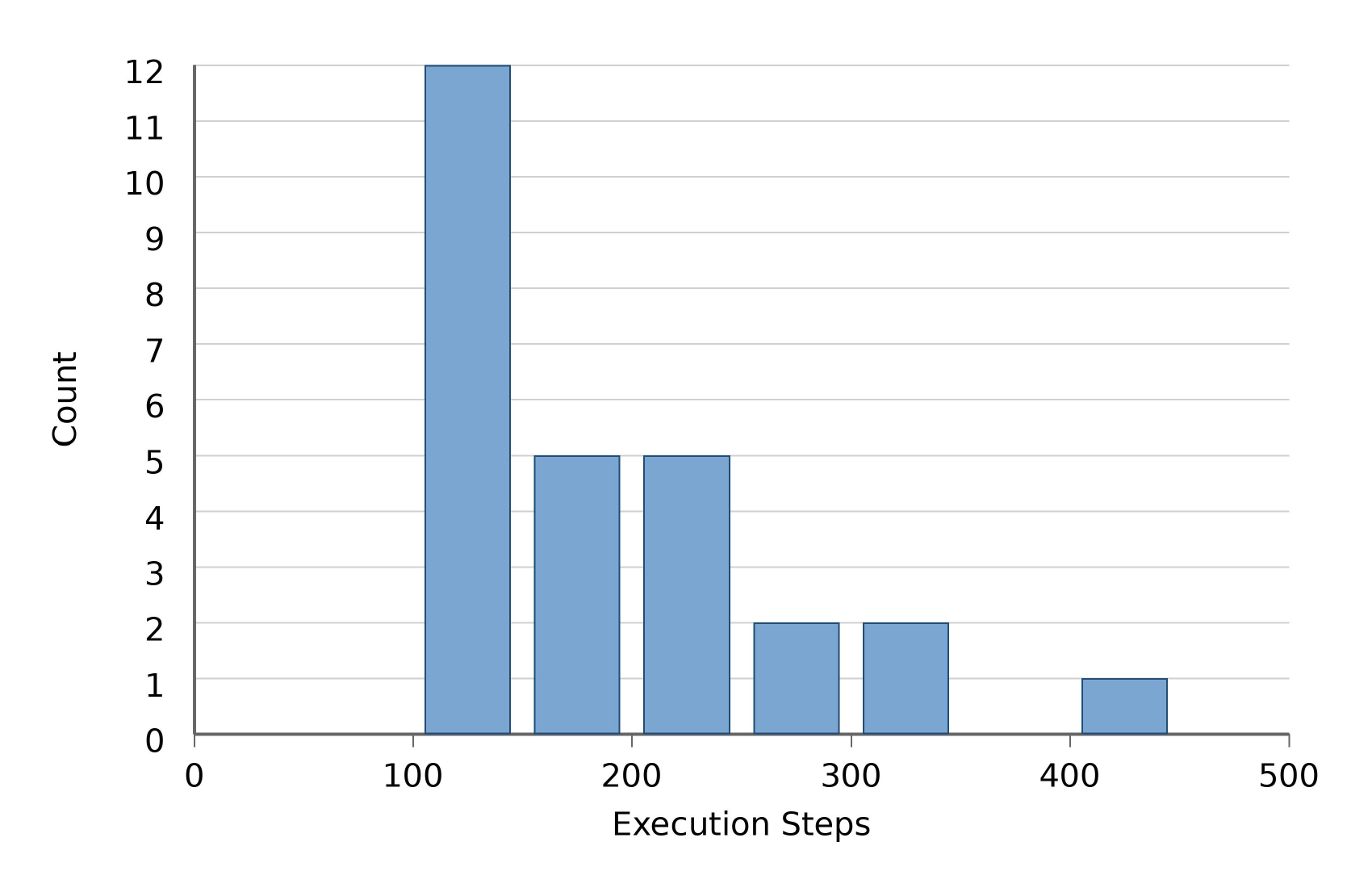}
        \caption{Kill Skeleton}
    \end{subfigure}\hfill
    \begin{subfigure}[t]{0.24\textwidth}
        \centering
        \includegraphics[width=\linewidth,trim=70 42 70 34,clip]{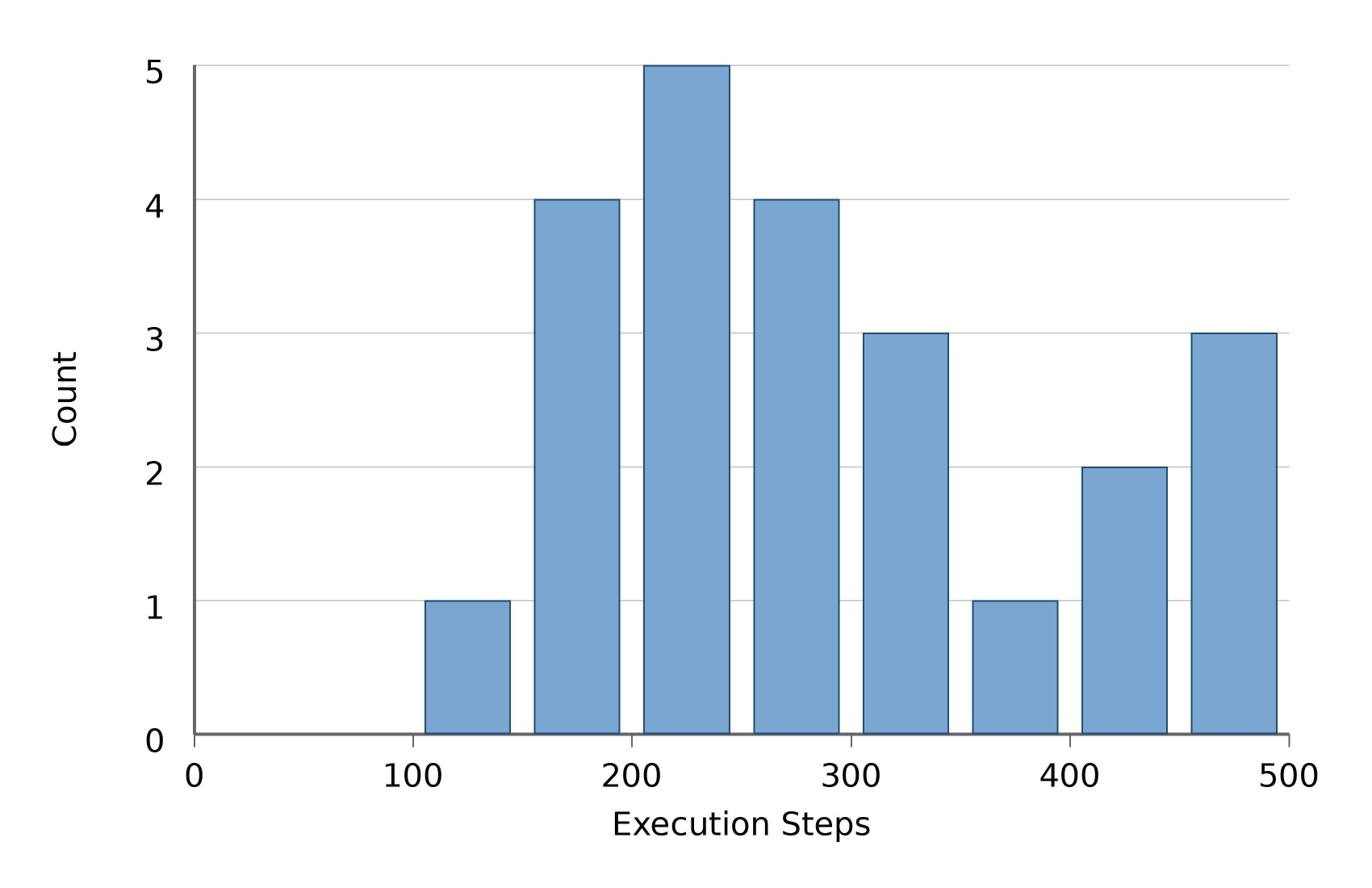}
        \caption{Kill Blaze}
    \end{subfigure}
    \vspace{-0.25em}
    \caption{Step count histograms over 30 rollouts for eight RL tasks.}
    \label{fig:appendix_step_histograms}
    \vspace{-1.25em}
\end{figure*}

Figure~\ref{fig:appendix_step_histograms} shows the step distribution across tasks. We observe that trajectories for the same task have similar numbers of execution steps. For failed tasks, the early termination stage also keeps trajectory lengths as close as possible to those of successful executions.

\section{Dataset Construction}
\label{appendix:dataset}
In our training framework, the cold-start stage is the only phase necessitating external data collection. Conversely, the next stage, multi-turn RL with GRPO, relies solely on predefined scenarios and resource configurations for each task, eliminating the need for additional data acquisition.

We first adopted the approach from JARVIS-VLA~\citep{li-etal-2025-jarvis} by utilizing a dataset rich in world knowledge for supervised fine-tuning. This component comprises 25k visual question answering samples, 225k grounding samples, and 9k captioning samples. Subsequently, we performed detailed processing on the open-source human player trajectory dataset from VPT~\citep{baker2022video} to construct an instruction-annotated action dataset.

Labeling and Segmentation We identified task completion based on transitions in the player's state recorded within the dataset. For instance, if the player's inventory showed the acquisition of an iron ore relative to the previous state, we labeled the preceding trajectory segment with the instruction to "mine iron ore." We defined the segment duration as the interval starting from the completion of the previous task to the moment the current state change occurred. This logic applies similarly to other interactions, such as labeling the interval preceding the defeat of a creature or the acquisition of a specific item with the corresponding objective.

Cleaning and Chunking Following the data cleaning protocol in VPT~\citep{baker2022video}, we filtered the dataset to remove redundant idle sequences. Specifically, if more than three consecutive idle frames were detected, we discarded the redundant frames starting from the fourth frame. To construct action chunks, we grouped actions into sets of four. For each set, we retained the observation solely from the first frame and concatenated the actions chronologically. We applied padding with null actions for groups containing fewer than four inputs.

Balancing and Final Composition An analysis of the dataset composition revealed a severe distributional imbalance, as the task of mining stone constituted 39.45\% of the total data. This skew significantly compromised the performance of the initial SFT model and hindered the subsequent RFT process. We addressed this issue by downsampling the overrepresented tasks to achieve a balanced distribution, which enabled the successful training of our supervised baseline. Through these processing and balancing steps, we ultimately collected approximately 4 million state-action samples for SFT.

\section{Hyperparameters}
\label{appendix:hyperparameters}
Our training framework consists of three progressive stages: (i) World Knowledge Learning, (ii) Imitation Learning, and (iii) Multi-Turn RL. The specific training parameters for all three stages are detailed in Table \ref{tab:hyperparameters}.

\begin{table}[H]
  \caption{Hyperparameter settings across different training stages.}
  \label{tab:hyperparameters}
  \centering
  \renewcommand\arraystretch{1.2}
  \begin{tabular}{@{}lccc@{}}
    \toprule
    \textbf{Hyperparameter} & \textbf{World Knowledge} & \textbf{Imitation Learning} & \textbf{Multi-Turn RL} \\ 
    \midrule
    Trainable Components & Full & Language Models & Full \\
    LR Scheduler & Cosine & Cosine & - \\
    Warm-up Ratio & 0.1 & 0.1 & - \\
    Global Batch Size & 32 & 32 & 8 \\
    Optimizer & AdamW & AdamW & AdamW \\
    Learning Rate & $1\times10^{-5}$ & $1\times10^{-5}$ & $1\times10^{-6}$ \\
    Group Size ($G$) & - & - & 8 \\
    Clipping Parameter ($\epsilon$) & - & - & 0.2 \\ 
    \bottomrule
  \end{tabular}
\end{table}

\section{Extending GROW to Simulated Language Table}
\label{appendix:lt_train}

To demonstrate that GROW remains effective in a markedly different embodied manipulation setting and is not limited to Minecraft, we retrain GROW on the simulated Language Table~\citep{lynch2022interactivelanguagetalkingrobots}, which provides a tabletop environment characterized by continuous control dynamics.

\subsection{Experimental Setup}
The policy receives an egocentric RGB view from a tabletop manipulation robot as input. Each state is represented by an image with resolution $180 \times 320 \times 3$. The action space consists of two-dimensional control values, corresponding to the displacement of the robot end effector along the $x$ and $y$ axes. To keep the action interface consistent with the Minecraft setting, we use $\mu$-law encoding to discretize the $x$ and $y$ axes independently into 21 bins, yielding 42 action tokens in total. Each bin is mapped to a reserved token.

For multi-turn RL, we follow the task setup of the simulated Language Table~\citep{lynch2022interactivelanguagetalkingrobots} and train on five task families: \texttt{block2block}, \texttt{block2abs}, \texttt{block2rel}, \texttt{block2blockrel}, and \texttt{separate}. The detailed definitions, success criteria, and the exact number of conditions for each task family are summarized in Table~\ref{tab:task_definitions}.

\begin{table}[htbp]
    \centering
    \caption{Definitions and configurations of the five task families in the simulated Language Table environment.}
    \label{tab:task_definitions}
    \begin{tabularx}{\textwidth}{@{} l X X l @{}}
        \toprule
        \textbf{Task Family} & \textbf{Agent Action} & \textbf{Success Criterion} & \textbf{\# Conditions} \\
        \midrule
        
        \texttt{block2block} 
        & Pushes a source block to another target block. 
        & Distance between the source and target blocks is below a threshold. 
        & 56 \footnotesize{(8 src $\times$ 7 tgt)} \\
        \addlinespace 
        
        \texttt{block2abs} 
        & Pushes a block to an absolute board location (9 locations\footnotemark[1]). 
        & Distance between the block and the target location is below a threshold. 
        & 72 \footnotesize{(8 blk $\times$ 9 loc)} \\
        \addlinespace
        
        \texttt{block2rel} 
        & Pushes a block to a relative offset location (8 directions\footnotemark[2]). 
        & Distance between the block and the invisible target offset location is below a threshold. 
        & 64 \footnotesize{(8 blk $\times$ 8 dir)} \\
        \addlinespace
        
        \texttt{block2blockrel} 
        & Pushes a source block to a relative offset location of another block (8 directions\footnotemark[2]). 
        & Distance between the source block and the invisible target offset of the target block is below a threshold. 
        & 448 \footnotesize{(8 src $\times$ 7 tgt $\times$ 8 dir)} \\
        \addlinespace
        
        \texttt{separate} 
        & Separates two blocks. 
        & Distance between the two blocks exceeds a predefined threshold. 
        & 56 \footnotesize{(8 src $\times$ 7 tgt)} \\
        
        \bottomrule
    \end{tabularx}
    
    \vspace{1ex}
    \raggedright
    \footnotesize{
    \textsuperscript{1} \textit{9 absolute locations:} top left, top center, top right, center left, center, center right, bottom left, bottom center, and bottom right. \\
    \textsuperscript{2} \textit{8 relative directions:} left, right, up, down, up-left, up-right, down-left, and down-right.
    }
\end{table}

\subsection{Implementation Details}
First, we initialize the policy of Qwen2.5-VL-7B-Instruct~\citep{bai2025qwen25vltechnicalreport} via SFT on 1M state-action samples. We construct the SFT dataset from the real-robot Language Table dataset by decomposing each episode into individual time steps. For each time step, the corresponding visual state and annotated control signal form one state-action sample. This procedure yields 1M state-action samples for policy initialization. We train the cold-start policy on 4 H200 GPUs for 8000 steps for about 3 days with a global batch size of 32. We use AdamW as the optimizer with a cosine learning-rate schedule and set the warm-up ratio to 0.1. During SFT, all model components are updated.

The simulated Language Table environment provides built-in task-specific verifiers for these task families. We use these verifiers to compute binary episode-level rewards according to the corresponding thresholded geometric success conditions. During RL, we train the policy with GROW for 200 training steps on 2 H200 GPUs for about 2 Days. The main hyperparameters are summarized in Table~\ref{tab:lt_hyperparameters}.

\begin{table}[H]
  \caption{Hyperparameter settings for policy initialization in simulated Language Table.}
  \label{tab:lt_hyperparameters}
  \centering
  \renewcommand\arraystretch{1.2}
  \begin{tabular}{@{}lc@{}}
    \toprule
    Hyperparameter & Value \\
    \midrule
    Global Batch Size & 16 \\
    Optimizer & AdamW \\
    Learning Rate & $1\times10^{-6}$ \\
    Group Size ($G$) & 8 \\
    Discount Factor ($\gamma$) & 0.995 \\
    Clipping Parameter ($\epsilon$) & 0.2 \\
    \bottomrule
  \end{tabular}
\end{table}

\subsection{Results}

As shown in Table~\ref{tab:lt_tasks}, the results show three main trends. First, GROW achieves the highest success rate across all five simulated Language Table task families, improving the average success rate from $65.02$ with PPO to $79.41$. Second, the gains are especially clear on spatial reasoning tasks: GROW exceeds PPO by $21.94$ points on \texttt{block2rel}, by $18.70$ points on \texttt{block2abs}, and improves \texttt{block2blockrel} from $46.91$ to $59.47$, suggesting stronger grounding of visual states, spatial targets, and action decisions. Third, the comparison with the initialized policy shows that RL is critical: the initialized policy only achieves $0.02$--$0.03$ on the four block-pushing tasks and $18.8$ on \texttt{separate}, while GROW reaches at least $59.47$ on all task families and $100.00$ on \texttt{separate}. These results indicate that GROW substantially improves closed-loop manipulation beyond supervised initialization and is effective in a continuous-control embodied environment.
\begin{table}[t]
\centering
\caption{Success rates on the simulated Language Table tasks.}
\label{tab:lt_tasks}
\begin{tabular}{lccccc}
\toprule
Method & \texttt{block2block} & \texttt{block2abs} & \texttt{block2rel} & \texttt{block2blockrel} & \texttt{separate} \\
\midrule
Initialized Policy & 0.02 & 0.03 & 0.02 & 0.02 & 18.8 \\
PPO                & 53.11 & 62.53 & 65.62 & 46.91 & 96.93 \\
\rowcolor{rowlightblue}
\textbf{Ours}               & \textbf{68.81} & \textbf{81.23} & \textbf{87.56} & \textbf{59.47} & \textbf{100.00} \\
\bottomrule
\end{tabular}
\end{table}


\section{Case Study in Minecraft}
To visualize the agent's decision-making, we overlay its action space on the left of each case study frame. We use a color-coded telemetry system to distinguish intent: yellow denotes inactive control primitives, while red signifies the specific actions selected by the model at that time step. This provides a direct, interpretable trace of the agent's behavioral primitives—such as movement keys or camera adjustments—relative to the visual context.
\subsection{Case Study: Kill Guardian}
Task Dynamics This case study highlights the model's ability to handle complex adversarial dynamics in Minecraft. The target, a Guardian, presents a unique challenge: it is a ranged attacker that maintains a tactical distance, retreating when the player approaches too closely while staying within its own offensive range. Since the agent is equipped only with a sword (a melee weapon), it must master a sophisticated behavioral loop involving active search, tactical approach, and precise engagement under evasive conditions.

\begin{figure*}[t]
    \centering
    \begin{subfigure}[t]{0.32\linewidth}
        \centering
        \includegraphics[width=\linewidth]{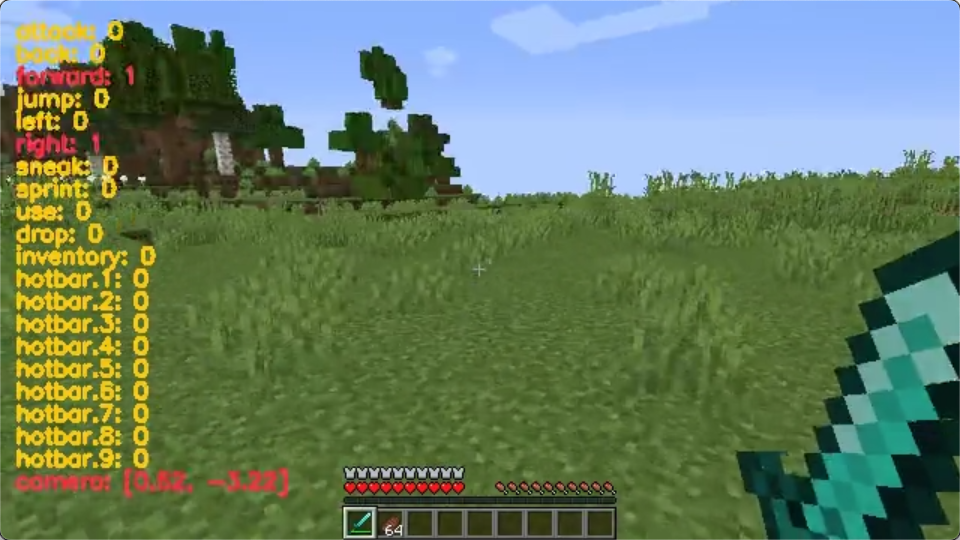}
        \caption{}
        \label{fig:guardian_search}
    \end{subfigure}\hfill
    \begin{subfigure}[t]{0.32\linewidth}
        \centering
        \includegraphics[width=\linewidth]{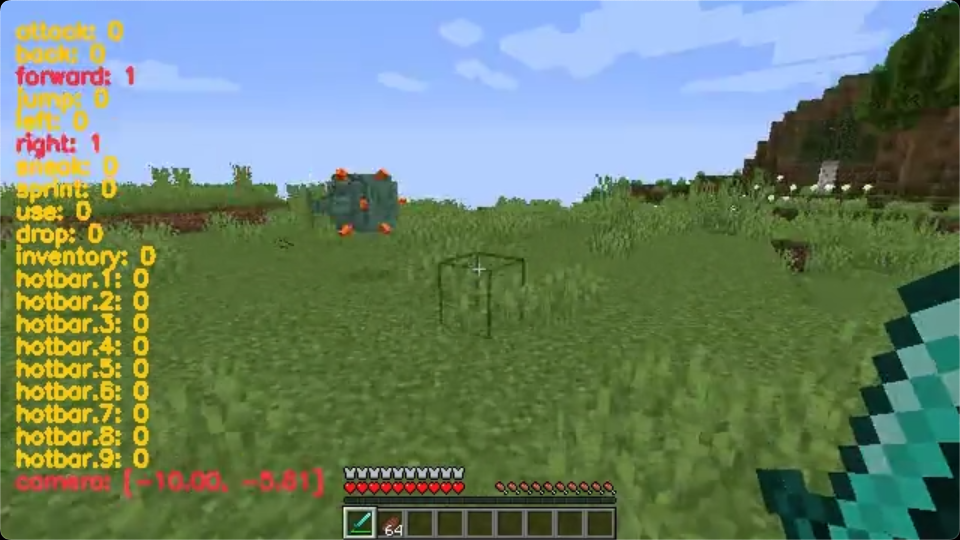}
        \caption{}
        \label{fig:guardian_search_b}
    \end{subfigure}\hfill
    \begin{subfigure}[t]{0.32\linewidth}
        \centering
        \includegraphics[width=\linewidth]{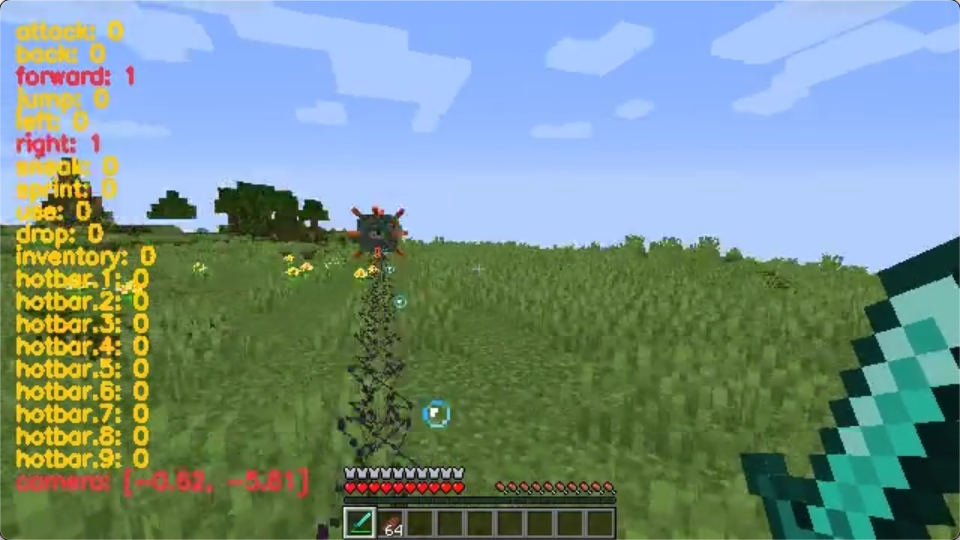}
        \caption{}
        \label{fig:guardian_evasion}
    \end{subfigure}

    \vspace{0.5em}

    \begin{subfigure}[t]{0.32\linewidth}
        \centering
        \includegraphics[width=\linewidth]{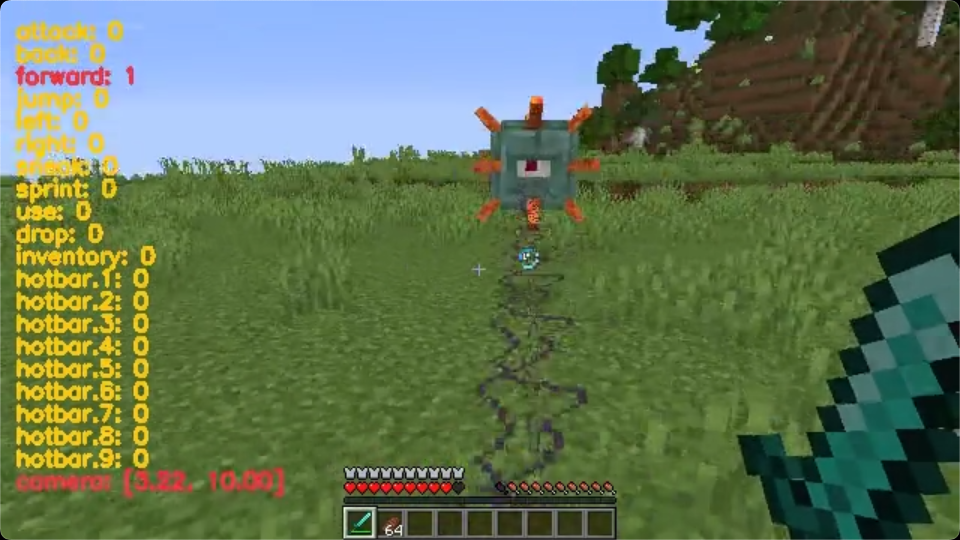}
        \caption{}
        \label{fig:guardian_evasion_b}
    \end{subfigure}\hfill
    \begin{subfigure}[t]{0.32\linewidth}
        \centering
        \includegraphics[width=\linewidth]{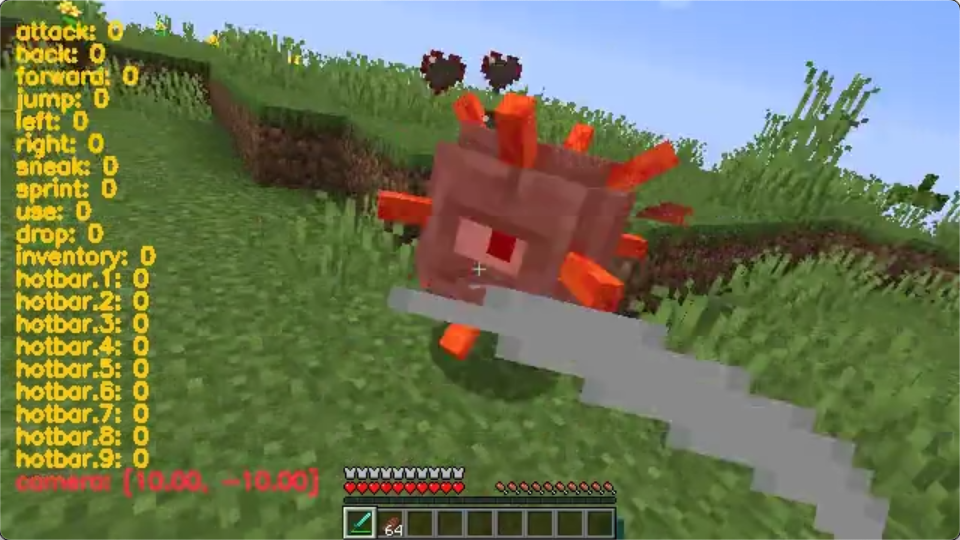}
        \caption{}
        \label{fig:guardian_melee}
    \end{subfigure}\hfill
    \begin{subfigure}[t]{0.32\linewidth}
        \centering
        \includegraphics[width=\linewidth]{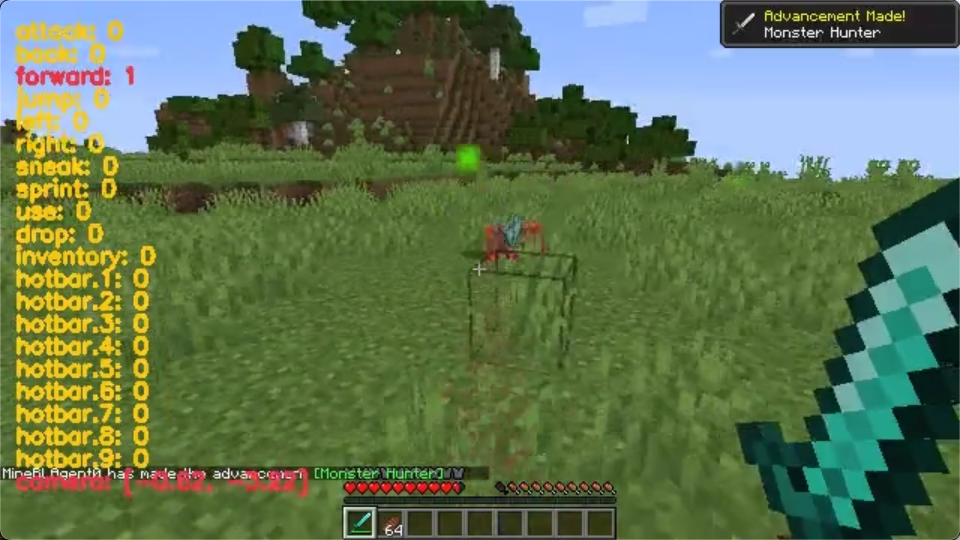}
        \caption{}
        \label{fig:guardian_melee_b}
    \end{subfigure}

    \caption{Behavioral analysis of the agent engaging a Guardian in combat.}
    \label{fig:kill_guardian}
\end{figure*}

Target Search (Figure \ref{fig:guardian_search},\ref{fig:guardian_search_b}): When the Guardian moves out of the field of view (FOV), the agent does not wander aimlessly. Instead, it performs a systematic search, utilizing camera yaw and pitch to scan the environment until the target is re-acquired.

Evasion and Approach (Figure \ref{fig:guardian_evasion},\ref{fig:guardian_evasion_b}): The agent identifies the Guardian’s ranged beam attack and initiates a direct approach. It successfully closes the gap despite the Guardian’s attempts to maintain distance.

Melee Engagement (Figure \ref{fig:guardian_melee},\ref{fig:guardian_melee_b}): Once within striking distance, the agent executes precise attack primitives while simultaneously tracking the target's movements, eventually neutralizing the threat through melee strikes.

\subsection{Case Study: Craft Stone Shovel}
Synthesizing advanced tools in Minecraft requires the agent to transition from spatial navigation to abstract GUI manipulation. In this case study, the agent must perform a multi-stage sequence: locating and interacting with a Crafting Table, navigating a multi-page recipe book to identify a specific target (Stone Shovel), and executing the final synthesis. This task tests the model's ability to maintain long-term goal coherence across different visual modalities (3D world vs. 2D interface).

\begin{figure}[htbp]
    \centering
    \begin{subfigure}[b]{0.24\linewidth}
        \includegraphics[width=\linewidth]{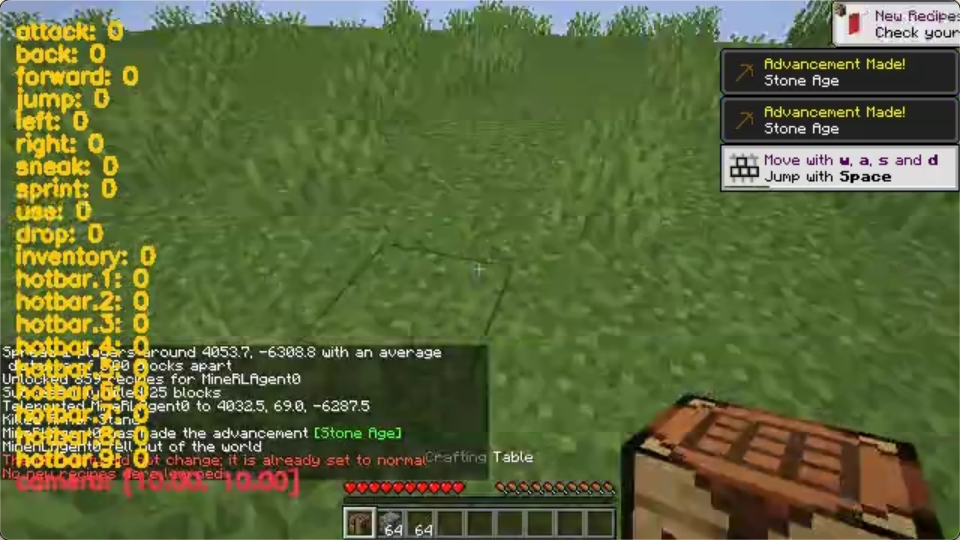}
        \caption{\label{fig:rollout_crafta}}
    \end{subfigure}
    \begin{subfigure}[b]{0.24\linewidth}
        \includegraphics[width=\linewidth]{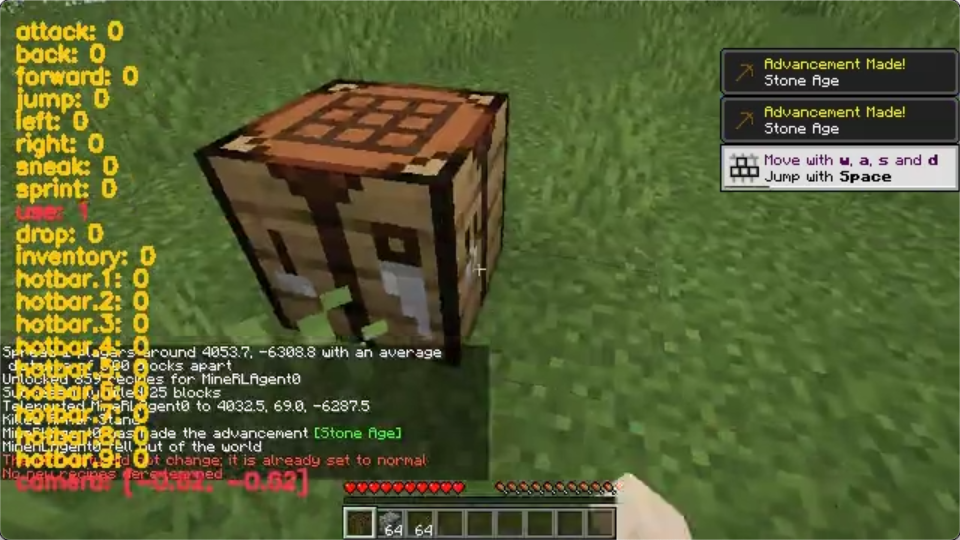}
        \caption{\label{fig:rollout_craftb}}
    \end{subfigure}
    \begin{subfigure}[b]{0.24\linewidth}
        \includegraphics[width=\linewidth]{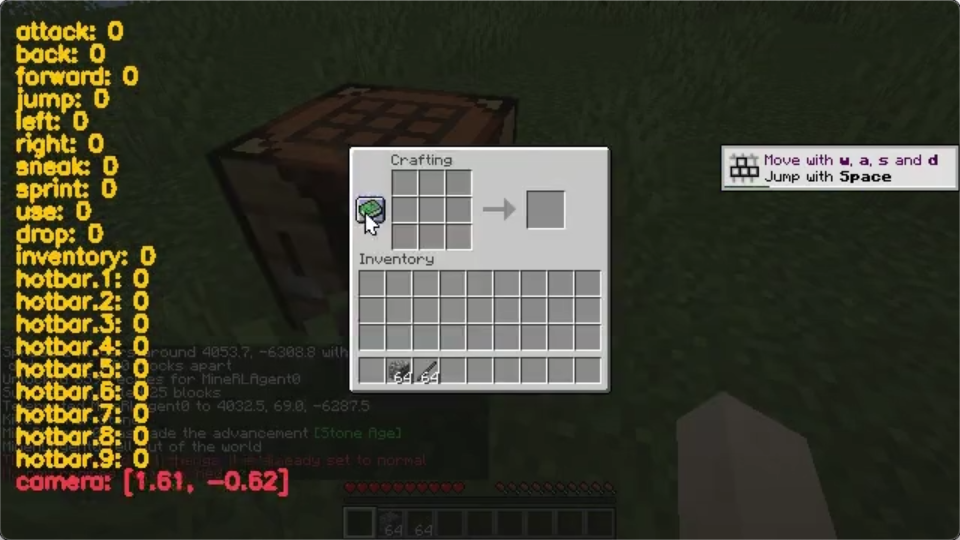}
        \caption{\label{fig:rollout_craftc}}
    \end{subfigure}
    \begin{subfigure}[b]{0.24\linewidth}
    \includegraphics[width=\linewidth]{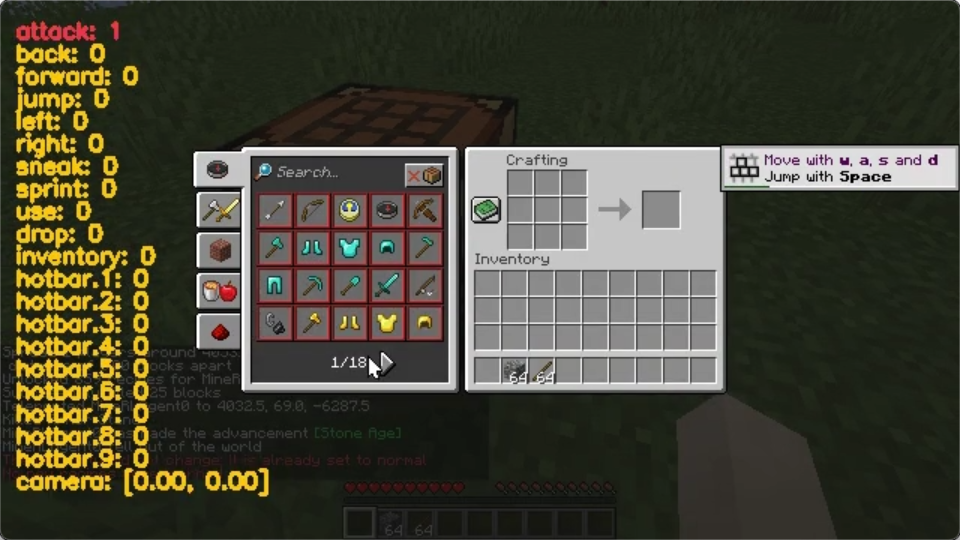}
        \caption{\label{fig:rollout_craftd}}
    \end{subfigure}
    \begin{subfigure}[b]{0.24\linewidth}
        \includegraphics[width=\linewidth]{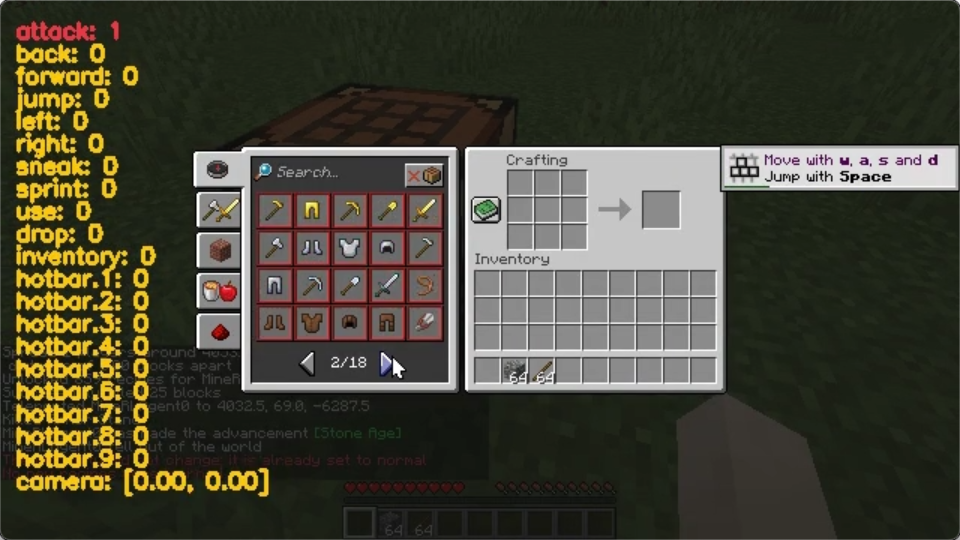}
        \caption{\label{fig:rollout_crafte}}
    \end{subfigure}
    \begin{subfigure}[b]{0.24\linewidth}
        \includegraphics[width=\linewidth]{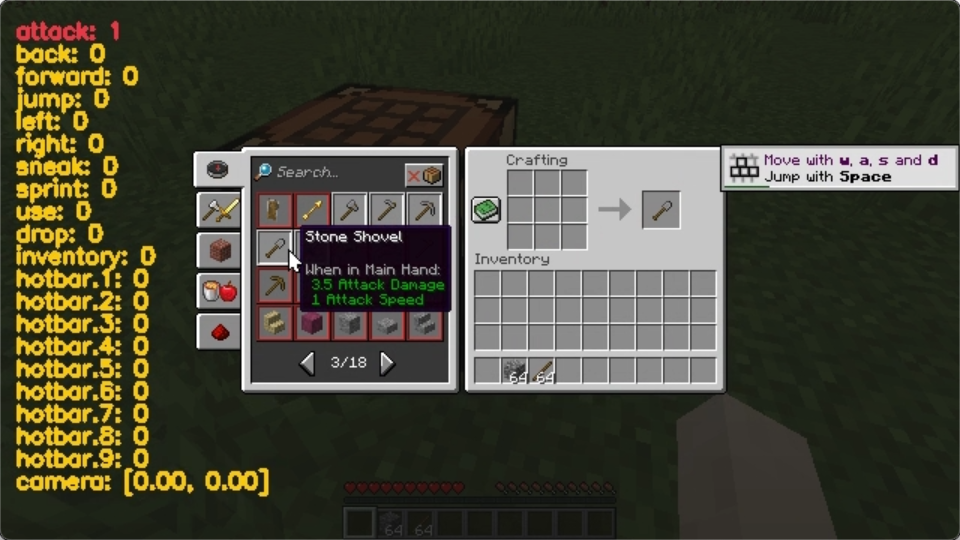}
        \caption{\label{fig:rollout_craftf}}
    \end{subfigure}
    \begin{subfigure}[b]{0.24\linewidth}
        \includegraphics[width=\linewidth]{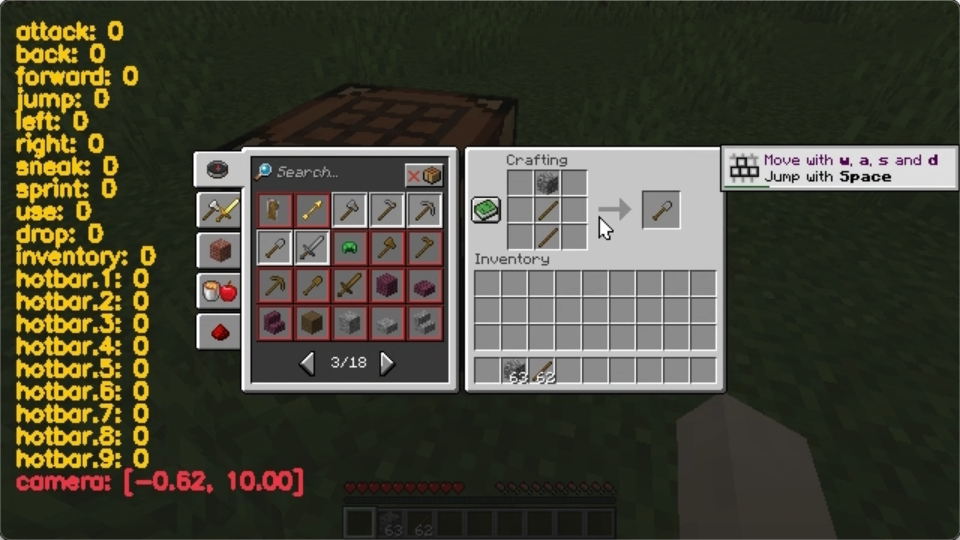}
        \caption{\label{fig:rollout_craftg}}
    \end{subfigure}
    \begin{subfigure}[b]{0.24\linewidth}
        \includegraphics[width=\linewidth]{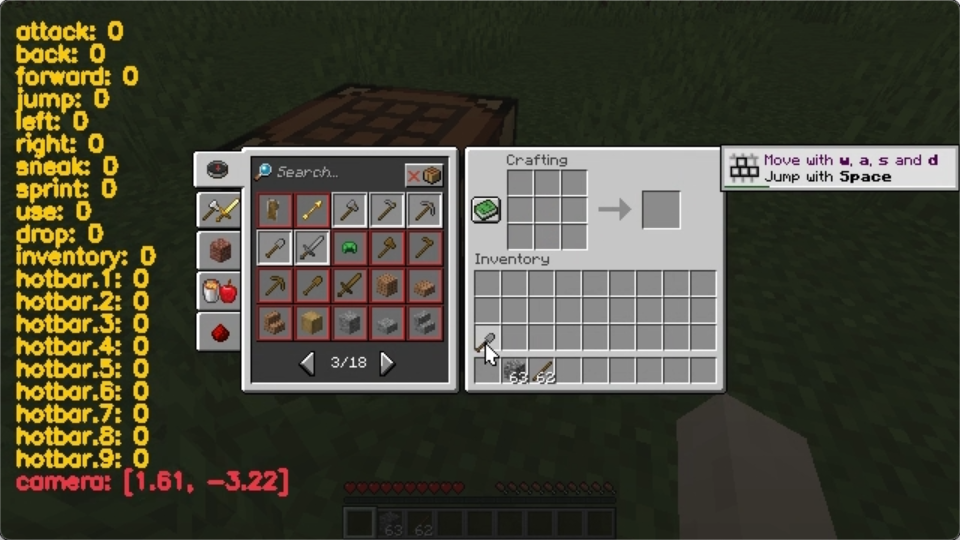}
        \caption{\label{fig:rollout_crafth}}
    \end{subfigure}
    \caption{Case study of the agent synthesizing a Stone Shovel.}
    \label{fig:rollout_craft}
\end{figure}

Real-time Action Telemetry As visualized in Figure \ref{fig:rollout_craft}, we use a color-coded telemetry overlay to trace the agent's intent: yellow denotes inactive controls, while red highlights active primitives. This allows us to observe the shift from movement-based exploration to precise UI-based clicking.

Environment Interaction (Figure \ref{fig:rollout_crafta},\ref{fig:rollout_craftb}): The agent approaches the Crafting Table and triggers the Use primitive (red) to open the synthesis interface.

GUI Navigation (Figure \ref{fig:rollout_craftc},\ref{fig:rollout_craftd},\ref{fig:rollout_crafte}): Once the interface is active, the agent’s focus shifts to the recipe book. The telemetry reveals a sequence of precise Attack (simulated as "click") and Camera (mouse cursor movement) primitives as the agent scrolls through multiple pages to locate the Stone Shovel recipe.

Target Synthesis (Figure \ref{fig:rollout_craftf},\ref{fig:rollout_craftg},\ref{fig:rollout_crafth}): Upon selecting the correct recipe, the agent executes the crafting command. The successful completion is confirmed by the appearance of the item in the output slot and the subsequent "Stone Age" advancement notification.

\section{Prompt in Experiments}
\begin{figure}[H] 
  \centering
  \includegraphics[width=\columnwidth]{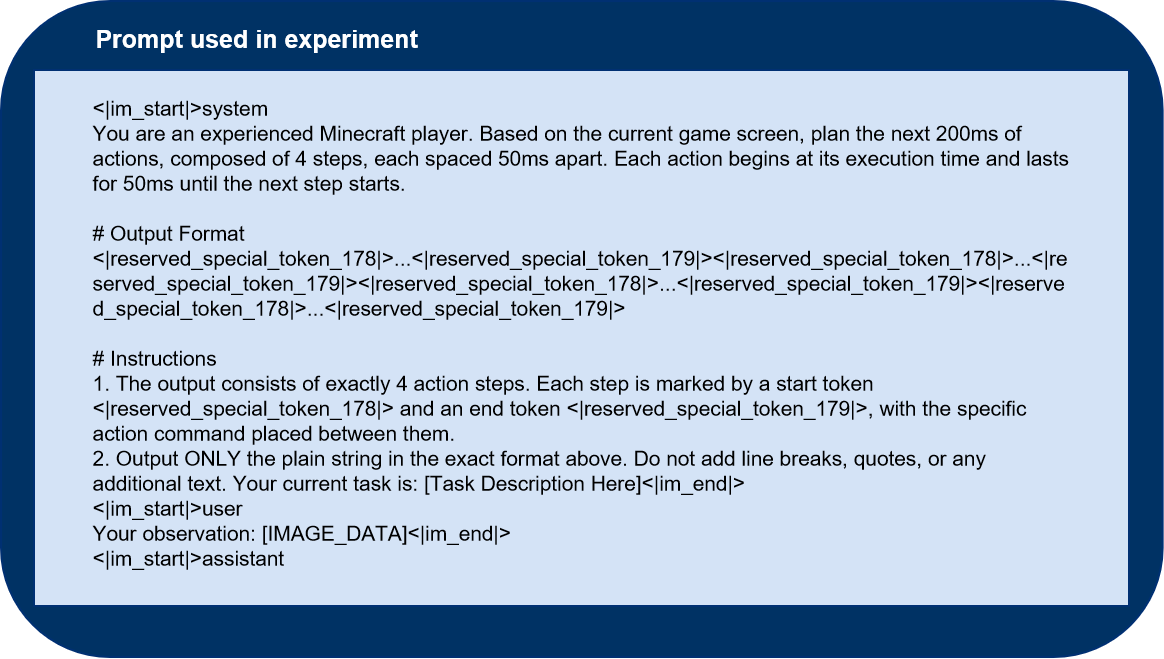} 
  \caption{}
  \label{fig:prompt}
\end{figure}

We show our prompt in Figure~\ref{fig:prompt}, which is used in our experiment.




\newpage

\end{document}